\documentclass[default,iicol]{sn-jnl}

\jyear{2024}
\theoremstyle{thmstyleone}

\theoremstyle{thmstyletwo}

\theoremstyle{thmstylethree}

\usepackage{booktabs}

\usepackage{amssymb}
\usepackage{booktabs}
\usepackage{subfig}

\usepackage{threeparttable}
\usepackage{multirow}
\usepackage{booktabs}
\usepackage{pifont}
\usepackage{graphicx}
\usepackage{amsmath}
\usepackage{array}
\usepackage{enumitem}

\usepackage{xcolor, soul}

\usepackage[capitalize]{cleveref}
\crefname{section}{Sec.}{Secs.}
\Crefname{section}{Section}{Sections}
\Crefname{table}{Table}{Tables}
\crefname{table}{Tab.}{Tabs.}

\def\ie{\emph{i.e.}}
\def\eg{\emph{e.g.}}

\def\etal{{\em et al.}}

\begin{document}

\title[Sparse Depth Completion]{\textbf{Towards Domain-agnostic Depth Completion}}

\author{
Guangkai Xu$ ^{1*}$,
~~~
Wei Yin$ ^{2*}$,
~~~
Jianming Zhang$ ^{3}$,
~~~
Oliver Wang$ ^{3}$, \\
~~~
Simon Niklaus$ ^3$
~~~
Simon Chen$ ^3$
~~~
Jia-Wang Bian$ ^{4\dag}$ \\

$^1$ Zhejiang University, China
~
$ ^2$ DJI Technology, China
~
$ ^3$ Adobe Research, United States
~
$ ^4$ University of Oxford, United Kindom \\
}

\abstract{
Existing depth completion methods are often targeted at a specific sparse depth type and generalize poorly across task domains.
We present a method to complete sparse/semi-dense, noisy, and potentially low-resolution depth maps obtained by various range sensors, including those in modern mobile phones, or by multi-view reconstruction algorithms.
Our method leverages a data-driven prior in the form of a single image depth prediction network trained on large-scale datasets, the output of which is used as an input to our model. 
We propose an effective training scheme where we simulate various sparsity patterns in typical task domains. In addition, we design two new benchmarks to evaluate the generalizability and robustness of depth completion methods.
Our simple method shows superior cross-domain generalization ability against state-of-the-art depth completion methods,  
introducing a practical solution to high-quality depth capture on a mobile device. 
}

\keywords{Monocular Depth Estimation, Depth Completion, Zero-shot Generalization, Scene Reconstruction, Neural Network}

\maketitle

\renewcommand{\thefootnote}{\fnsymbol{footnote}}

\footnotetext[1]{Both authors contributed equally. $^\dag$ Corresponding author. E-mail: jiawang.bian@gmail.com}

\section{Introduction}
Accurate metric depth is important for many computer vision applications, in particular 3D perception~\cite{shi2020pv,wang2019pseudo} and reconstruction~\cite{newcombe2011kinectfusion,murORB2,xu2023frozenrecon}.
Typically, depth is obtained by using direct range sensors such as LiDAR and Time-of-Flight (ToF) sensors on modern mobile phones or computed by multi-view stereo methods~\cite{schops2017multi, yao2020blendedmvs}.
However, usually, these sources can only provide incomplete and/or sparse depth information.
For example, LiDAR sensors capture depth in a linear scanning pattern, ToF sensors are lower resolution and fail at specular or distant surfaces, and multi-view reconstruction methods~\cite{schops2017multi, yao2020blendedmvs,Zhang2019GANet} only provide confident depth at textured regions and are range limited by the camera baseline (the iPhone rear stereo camera has a maximum depth of $2.5$ meters).

\begin{figure*}[t]
\centering
\includegraphics[width=1\textwidth]{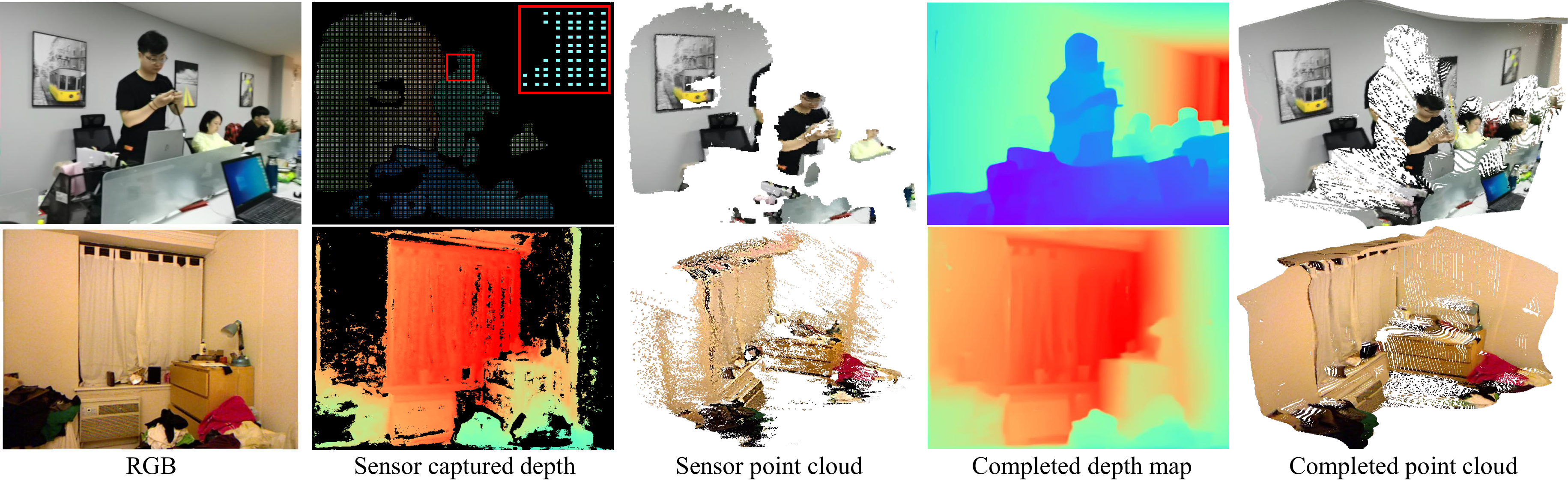}
\caption{Our method fills in missing information in different types of sparse depth maps. A single model can be used to complete the sparse depth from different methods, \eg Huawei Mate30 Time-of-Flight sensor (top) and a multi-view stereo algorithm~\cite{schops2017multi} (bottom).
}\label{Fig: first page fig.}
\end{figure*}

Existing depth completion methods can be classified into two categories according to the input sparsity pattern: depth inpainting methods that fill large holes~\cite{zhang2018deepdepth, senushkin2020decoder, huang2019indoor}, and sparse depth densification methods that densify sparsely distributed depth measurements~\cite{cheng2020cspn++, park2020non, xu2019depth, qiu2019deeplidar, cheng2019learning}. 
When working on a specific sparsity pattern, e.g., on either NYU~\cite{silberman2012indoor} or KITTI~\cite{Uhrig2017THREEDV}, 
recent approaches~\cite{park2020non, cheng2020cspn++, cheng2018depth, qiu2019deeplidar,imran2021depth} can obtain impressive performance. 
However, in real-world scenarios, the sparsity pattern may be subject to change or unknown at training time, as it is a function of hardware, software, as well as the configuration of the scene itself. 

In this paper, we revisit existing methods and analyze the performance gap between the training setup and downstream applications, 
and we find that existing depth completion methods suffer from the following key limitations. 
First, their methods work best on one specific sparsity pattern, but they are sensitive to mild perturbations of the depth sampling and generalize poorly to other types of sparse depth (e.g., from Kinect depths to smartphone depths).
Second, they are sensitive to noise and outliers produced by the depth capture process.
To address these issues, we propose a simple yet effective method for robust cross-domain depth completion. 

Our method provides the following improvements. 
First, inspired by domain randomization methods~\cite{tobin2017domain, tobin2018domain, zakharov2019deceptionnet}, we analyze the existing set of common sparsity patterns and create a diverse 
set of synthetic sparsity patterns to train our model.
To improve the cross-domain generalization ability, we follow recent monocular depth prediction methods~\cite{Wei2021CVPR, Ranftl2020} and utilize a diverse training dataset that consists of multiple depth sources.
Furthermore, to make our method robust to noise, we leverage the depth map predicted by a well-trained single image depth prediction method as a data-driven scene prior.
Such approaches learn a strong prior on diverse scenes~\cite{Ranftl2020}, but their predicted depths have unknown shifts and scales due to the training data used (stereo images with unknown baseline). 
By incorporating sparse metric depth cues and a single image relative depth prior, our method is able to robustly produce a dense metric depth map.

Furthermore, as current benchmarks on NYU~\cite{silberman2012indoor} and Matterport3D~\cite{Matterport3D} cannot evaluate a method's robustness to different sparsity types, noises, and diverse scenes, we redesign a benchmark based on existing datasets.
We combine 3 datasets, i.e. NYU~\cite{silberman2012indoor}, ScanNet~\cite{dai2017scannet}, and DIODE~\cite{vasiljevic2019diode}, which include both indoor and outdoor scenes.
We simulate 3 different sparsity types for model evaluation based on typical application scenarios.
Furthermore, we also set up another benchmark to evaluate a model's robustness to noisy sparse depth inputs, which can be produced by SLAM, SfM methods, or multi-view stereo methods. 
Therefore, we collect 16 NYU videos and employ COLMAP~\cite{schoenberger2016mvs} to create noisy sparse depth for completion. 

Our simple approach sets up a strong baseline on our new benchmarks for robust domain-agnostic depth completion. We also show that our method can generalize well on sparse depth captured by various smartphones, providing a practical solution to high-quality depth sensing on low-cost mobile devices.

In conclusion, our main contributions are as follows.
\begin{itemize}[noitemsep]
    \item We analyze existing ``RGB + sparse depth'' depth completion methods in terms of generalization across different domains and robustness to noise. 
    \item We propose a method that incorporates a data-driven single-image prior and effective data augmentation techniques for domain-agnostic depth completion. 
    \item We design two new synthetic benchmarks for evaluating the robustness and the generalizability of a depth completion method. Our new benchmark design simulates challenges in various real-world scenarios.
\end{itemize}
We hope that our new paradigm of training and benchmarking depth completion models can also evoke new thoughts on improving single depth estimation methods using sparse depth measurements\cite{huynh2021boosting, Ranftl2020,Wei2021CVPR}.

\section{Related Work}
\noindent\textbf{Sparse Depth Completion.}
Depth completion~\cite{imran2021depth, chen2019learningJoint} aims to densify a sparse depth input. 
As the sparse depths captured by different solutions have varying sparsity patterns, several methods are proposed to solve these problems. 
Depth maps captured with low-cost LiDAR only have a few hundred or thousand measurements per image.
Several methods~\cite{cheng2020cspn++, park2020non, cheng2019learning, chen2019learning, yang2019dense} propose to leverage texture information to complete these types of sparsity patterns.
Besides such very sparse depth types, commodity-level RGB-D cameras such as Kinect, RealSense, and Tango produce depth images that are semi-dense but missing certain regions.
This often happens due to objects with low reflective properties and objects beyond the maximum supported distance.
Several methods treat this as a depth inpainting task and leverage smoothness priors~\cite{herrera2013depth}, background surface extrapolation~\cite{matsuo2015depth}, and surface normals~\cite{zhang2018deepdepth}. 
These methods have shown promising results, but they focus on only a single sparse depth type. 
In contrast, we design a unified solution for all these depth sparsity patterns. 
Different from previous depth completion methods that use an RGB image and a sparse depth map as input, we propose to additionally leverage a pretrained scale-shift-invariant depth prediction model as a scene prior to improve the depth completion quality and robustness.

Our work mainly focuses on the paradigm of depth completion given RGB image and sparse depth.  Some existing works also deal with the depth-only completion setting. CU-Net~\cite{albishri2019cu} concentrates on the LiDAR depth-only completion task, BoostingDepth~\cite{xu2022towards} uses the locally weighted linear regression (LWLR) method to align the predicted depth with the ground-truth depth. For this depth-only setting, we can use a neural network to fit the transformation between sparse depth and completed depth without RGB information, \textit{i.e.}, reusing our guidance-map pipeline without inputting RGB image. The guidance map can offer structure information, and the sparse ground-truth depth can offer accurate adjustment direction of depth maps, and we think it will perform better than the simple LWLR of BoostingDepth. However, for this work, we only concentrate on the paradigm given both RGB image and sparse depth.

\noindent\textbf{Monocular Depth Estimation.}
Monocular depth estimation aims to  predict the depth from a single RGB image.
Eigen~\etal~\cite{eigen2014depth} proposed the first multi-scale neural network to predict depth from a single image.
Subsequently, many supervised~\cite{Yin2019enforcing, liu2015learning, xian2020structure} and self-supervised methods~\cite{bian2021ijcv,monodepth2} have been proposed to address this task. 
One problem is the limited availability of training data (especially metric depth), so recent methods~\cite{Ranftl2020, yin2020diversedepth, yin2020learning, ke2023repurposing, xu2024diffusion} that learn affine-invariant depth have shown that much better generalization properties can be obtained by training on diverse large-scale datasets that include data with an unknown shift and scale (e.g., from stereo images).
The affine-invariant depth estimation models are very popular in many applications such as offering geometric priors for neural reconstruction ~\cite{wang2022neuris, yu2023improving}. However, the unknown scale and shift of the predicted depth maps can bring duplication and distortion of the unprotected point clouds, as mentioned in BoostingDepth~\cite{xu2022towards}. These methods rely on performing least square fitting with ground-truth depth for practical application, which results in poor performance without considering the distribution difference between the predicted depth and ground-truth depth. One of our contributions is to propose a novel paradigm and learn to deal with the distribution difference inherently.

\begin{figure*}[t]
\centering
\includegraphics[width=1\linewidth]{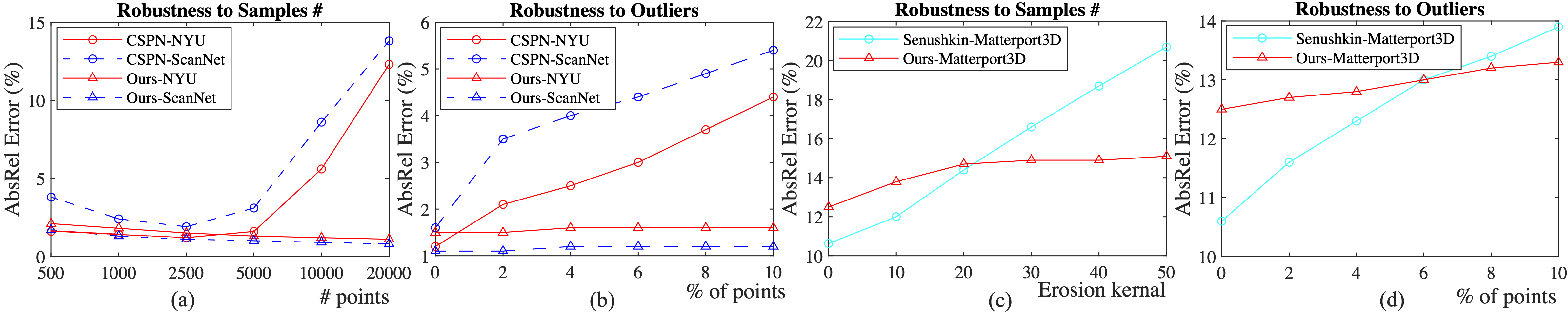}
\caption{Robustness analysis. We analyze the performance of CSPN~\cite{cheng2020cspn++} (completion) and Senushkin~\etal~\cite{senushkin2020decoder} (inpainting) in terms of input point numbers/patterns (a, c) and outlier ratios (b, d). CSPN is trained on NYU~\cite{silberman2012indoor}, and we evaluate it on both NYU and ScanNet~\cite{dai2017scannet}. Senushkin~\etal is trained and evaluated on Matterport3D~\cite{Matterport3D}.
\label{Fig: Robustness cmp.}
}
\end{figure*}

\section{Analysis of Existing Methods}

In this section, we evaluate two state-of-the-art depth completion methods to assess their performance with respect to different sparsity patterns, dataset generalization, and robustness to noise.
To do this, we perturb the sparsity pattern of the input depth in various ways and add additional noise to it.
We also evaluate methods on their zero-shot cross-dataset generalization performance~\cite{Ranftl2020} (evaluating on a different dataset than what the models were trained on).
We choose two methods for this analysis, CSPN~\cite{cheng2019learning} and Senushkin~\etal~\cite{senushkin2020decoder}.
The former is designed to complete very sparse depth with only hundreds of sparse points, while the latter is designed to complete contiguous holes.
We use the code and the model weights provided by the authors for this evaluation. As a reference, we also report the performance of our method, which is going to be presented in Sec.~\ref{sec:robust}. Note that our method is not trained on any of the datasets mentioned below.

For CSPN~\cite{cheng2019learning} which is trained on NYU~\cite{silberman2012indoor}, we use NYU and ScanNet~\cite{dai2017scannet} for testing, and
we vary the number of measured/input points from $500$ to $20000$.
Senushkin~\etal~\cite{senushkin2020decoder} is trained on Matterport3D with the task of completing holes of depth maps.
We use Matterport 3D for testing, and erode the valid depth regions with different kernel sizes to control the number of valid points on Matterport3D~\cite{Matterport3D}. 
Note that these perturbations do not significantly change the sparsity patterns, and a robust model should be able to handle these mild variations. 
However, from Fig.~\ref{Fig: Robustness cmp.} (a) and (c), we observe that their methods are sensitive to such mild perturbations. 
It is especially quite counterintuitive that the performance of CSPN degrades with the increase of the input points, showing that it generalizes poorly beyond the sampling density it was trained on. The depth completion method CSPN is trained on the NYU dataset with the sparse depth containing 500 points randomly sampled from the gt depth map following the previous work\cite{ma2018sparse}. Due to the lack of diversity of the input sparse depth paradigms, it shows less robustness when the number of samples increases. Different from CSPN, we simulate three various sparsity patterns, and randomize the parameter of sample points number, which improves the robustness significantly.

Besides, as outliers are unavoidable in many applications, we also simulate depth noise for both methods by sampling $0\% - 10\%$ points from the sparse depth and multiplying the original depth with a random factor from $0.1-2$. Fig.~\ref{Fig: Robustness cmp.} (b) and (d) show that their performance also degrades more significantly with the increase in outliers than our method.

\begin{table}[!t]
\centering
\footnotesize
\begin{tabular}{l|lll}
\toprule
\multirow{2}{*}{\begin{tabular}[c]{@{}l@{}}Sparsity \\ type 1\end{tabular}} & \multicolumn{3}{l}{Uniform (2500 points, AbsRel (\%)$\downarrow$)}   \\ \cline{2-4} 
                                                                          & NYU           & ScanNet          & Matterport3D        \\ \midrule
CSPN~\cite{cheng2018depth}                                                                     & \textbf{1.1}           & 2.0              & 27.9                \\
Senushkin~\etal~\cite{senushkin2020decoder}   & 77.3          & 62.1             & 69.5                \\ 
Ours      & \underline{1.5}  & \textbf{1.1}   & \textbf{2.2}                \\ \hline \hline
\multirow{2}{*}{\begin{tabular}[c]{@{}l@{}}Sparsity \\ type 2\end{tabular}} & \multicolumn{3}{l}{Features (2500 points, AbsRel(\%)$\downarrow$)} \\ \cline{2-4} 
                                                                          & NYU           & ScanNet          & Matterport3D        \\ \midrule
CSPN~\cite{cheng2018depth}         & 1.9           & 13.8             & 51.9                \\
Senushkin~\etal~\cite{senushkin2020decoder}     & 76.9          & 60.8             & 66.7                \\
Ours    & \textbf{1.5}   & \textbf{1.6}          & \textbf{2.7}               \\
\bottomrule
\end{tabular}
\caption{Robustness to different sparse depth patterns (AbsRel error). CSPN is trained on NYU to complete very sparse depth, while Senushkin~\etal~\cite{senushkin2020decoder} is trained on Matterport3D to complete holes. They are tested on 3 datasets with two different sparse depth types. The sparse depth of the `Uniform' type is sampled uniformly from the ground truth depth, while `Features' are sampled from FAST feature points.  Both methods are sensitive to the sparsity pattern and dataset domain. 
\label{Tab: Robustness cmp. }}
\end{table}

Furthermore, we test their robustness to different sparse depth patterns and cross-domains generalization (Table~\ref{Tab: Robustness cmp. }.
For both methods, we input two kinds of sparse depth, \ie uniform sparse depth (Uniform) and sparse depth based on points that are detected by the FAST~\cite{rosten2006machine} feature detector (Features). 
It shows that CSPN, which was trained on the uniform sparse depth, is sensitive to the point distribution (performance degrades a lot from `Uniform' to `Features'), while Senushkin~\etal~\cite{senushkin2020decoder} fails on both sparsity patterns. 
Besides, we can observe that CSPN is sensitive to the dataset domains. 
It performs much better on the training dataset (NYU) than the other two datasets. 
Results are summarized in Table~\ref{Tab: Robustness cmp. }. 
By contrast, our method is more robust to cross-domains and different sparsity patterns.

\begin{figure*}[t]
\centering
\includegraphics[width=\linewidth]{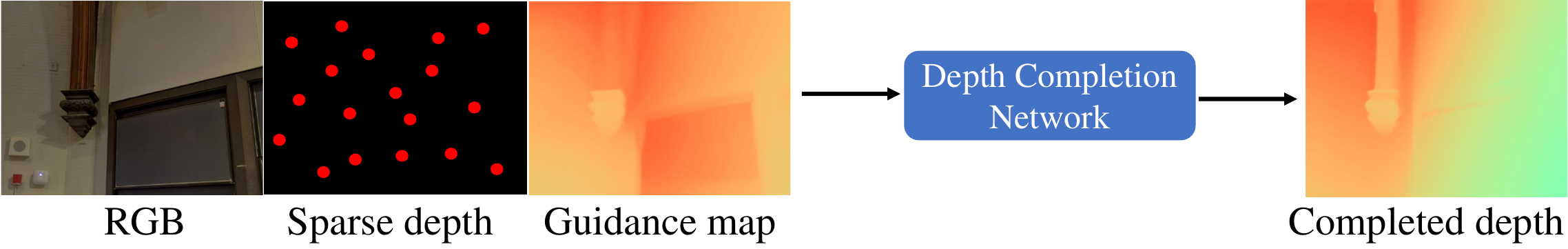}
\caption{Our method takes an RGB image, sparse depth, and guidance map as input, and it outputs a dense completed depth.
}
\label{Fig: refinement framework.}
\end{figure*}

\begin{figure}[!t]
\centering
\subfloat[]
{
	\includegraphics[trim=0cm 1.48cm 0cm 1.48cm, clip=true, width=0.45\linewidth]{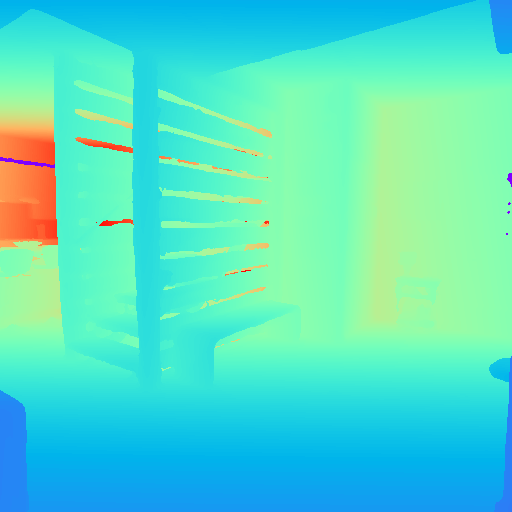}
	\label{Fig: original depth.}
}
\subfloat[]
{
	\includegraphics[trim=0cm 2cm 0cm 2cm, clip=true, width=0.45\linewidth]{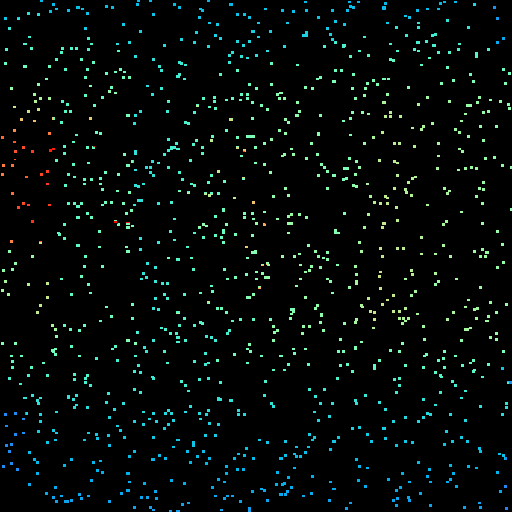}
	\label{Fig: uniform samples.}
}\\
\subfloat[]
{ 
	\includegraphics[trim=0cm 2cm 0cm 2cm, clip=true, width=0.45\linewidth]{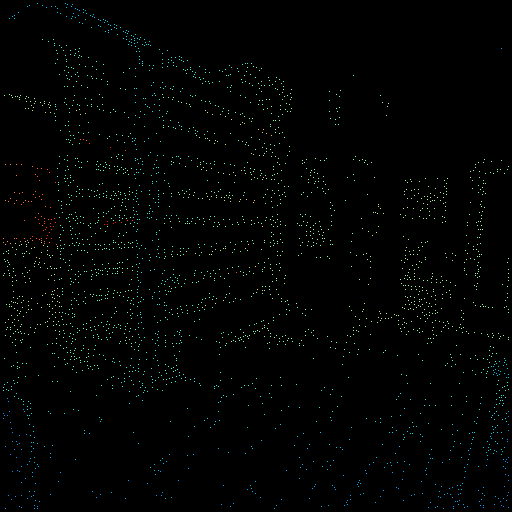} 
	\label{Fig: features.}
}
\subfloat[]
{ 
	\includegraphics[trim=0cm 2cm 0cm 2cm, clip=true, width=0.45\linewidth]{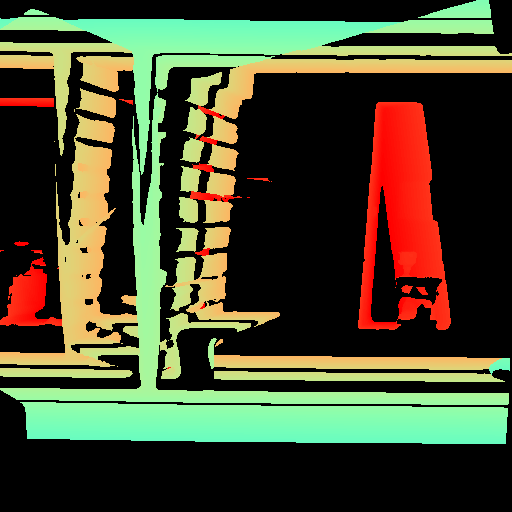} 
	\label{Fig: simulate sensor.}
}
\caption{Visualization of sampled sparse depths. We simulate three different patterns from (a) the dense depth for training models: (b) random uniform sampling, (c) feature point based sampling, and (d) region-based sampling.}
\label{Fig: sdepth patterns}
\end{figure}

\section{Robust Depth Completion}
\label{sec:robust}

We now introduce our approach, a sparse depth completion method that is designed to be robust to noise, applicable to different types of sparse depth, and to generalize well to unseen datasets.
\subsection{Model Architecture and Training}
The framework is illustrated in Fig.~\ref{Fig: refinement framework.}.
Our depth completion model takes as input the RGB image, sparse depth, as well as a guidance depth map, and it outputs a dense completed depth map. 
The sparse depth map supplies depth metric information, the guidance map offers robust geometric information, and they are combined implicitly with the help of our depth completion network, which performs much better than the traditional least square aligning. (See Table~\ref{Tab: DepComplete Cmp with SOTA on NYU}.)
We use the slightly modified ESANet-R34-NBt1D network with the ResNet-34 backbone proposed by Seichter~\etal~\cite{esanet2020} for depth completion, and we use the affine-invariant depth predicted by the method from Yin~\etal~\cite{Wei2021CVPR} as the guidance depth map. Concretely, the original ESANet~\cite{esanet2020} fuses the RGB branch and the depth branch to estimate semantic information. During training our robust depth completion model, we simply concatenate the sparse depth map, the valid mask of the sparse depth, and the predicted guidance map together, and regard it as the depth branch input of the ESANet.
We employ the virtual normal loss~\cite{Yin2019enforcing}, pair-wise normal regression loss~\cite{Wei2021CVPR}, ranking edge loss~\cite{xian2020structure}, and L1 loss to supervise training. 
The overall loss is defined as follows.
\begin{equation}
    L = L_{vnl} + L_{pwn} + L_{rel} + L_1
\end{equation}
We sample $36000$ images from Taskonomy~\cite{zamir2018taskonomy}, DIML~\cite{kim2018deep}, and TartainAir~\cite{wang2020tartanair} as the training data.

\subsection{Sparsity Pattern Generation} 
As we cannot access enough data to cover the diverse sparsity patterns of all possible downstream applications during training, we instead simulate a set of various sparsity patterns.
This approach is motivated by domain randomization methods~\cite{tobin2017domain, tobin2018domain, zakharov2019deceptionnet} that train models on simulated data and show that the domain gap to real data can be reduced by randomizing the rendering in the simulator.

We categorize the sparse depth patterns into three main classes, which are illustrated in Fig.~\ref{Fig: sdepth patterns}.
During training, we sample sparse points from the dense ground truth depth and try to recover the dense depth map.

\begin{itemize}[noitemsep]
    \item \textbf{Uniform.} We sample uniformly distributed points, from hundreds to thousands of points, to simulate the sparsity pattern from the low-resolution depths, e.g., those captured by ToF sensors on mobile phones.
    \item \textbf{Features.} In order to simulate the sparsity pattern from structure-from-motion and multi-view stereo methods, where high confidence depth values are produced only at regions with distinct/matchable features, we apply the FAST~\cite{rosten2006machine} feature detector that samples points on textured regions and particularly image corners.
    \item \textbf{Holes.} Commodity-grade depth sensors cannot capture depth on bright, transparent, reflective and distant surfaces. Therefore, multiple large coherent regions may be missing. 
    We simulate this by 1) masking the depth using a random polygonal region, 2) masking regions at a certain distance, and 3) masking the whole image with the exception of a polygonal region. 
\end{itemize}

To increase the diversity of these patterns, we augment each type of sparse depth by randomizing its parameters (\eg, the number of valid points, mask size, and feature thresholds), and then by combining sparsity patterns together. 

\noindent\textbf{Improving the Robustness to Outliers.}
Outliers and depth sensor noise are unavoidable in any depth acquisition method. 
Most previous methods only take an RGB image and a sparse depth as the input, and they do not have any extra source of information with which they could distinguish the outliers.
However, our method leverages a data prior from the single image depth network, which can help resolve incorrect constraints when there is a significant discrepancy between the two.
In order to encourage the network to learn this, we add outliers during training.
Specifically, we randomly sample several points and scale their depth by a random factor from $0.1$ to $2$.

\begin{figure*}[t]
\centering
\includegraphics[width=\linewidth]{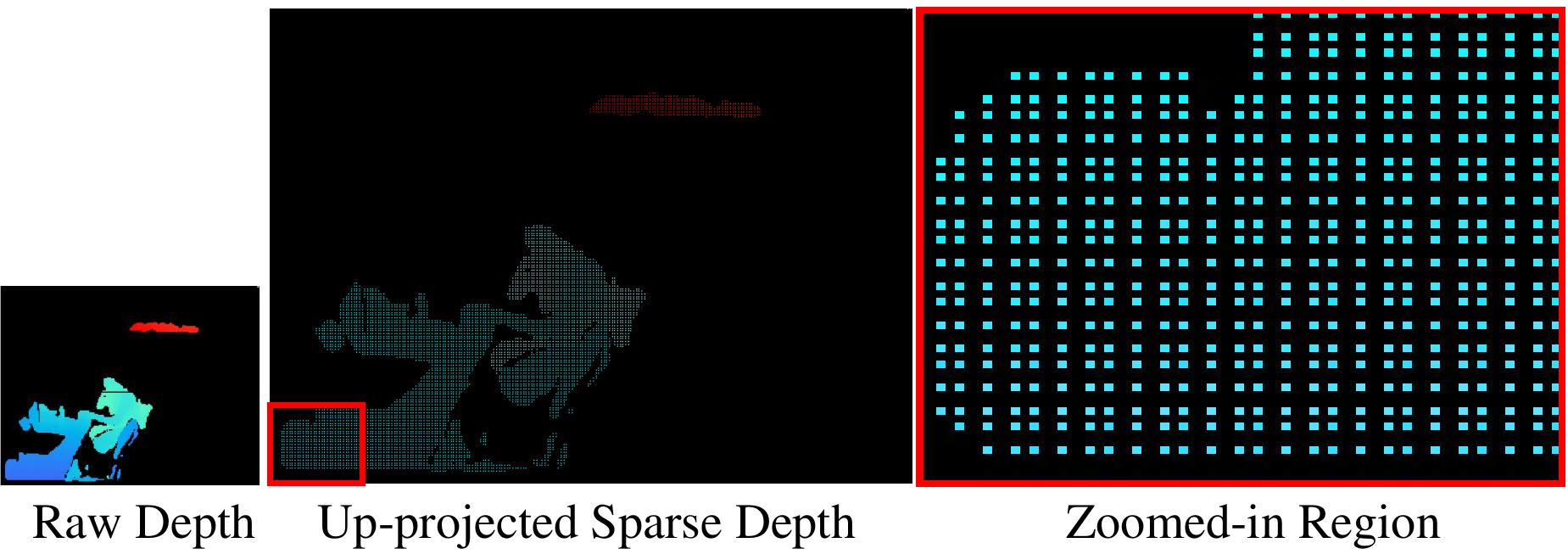}
\caption{RGBD capture using a Huawei phone and its up-projected sparse depth map. The depth and the RGB sensor have the same field of view but the resolution is different. RGB resolution is in $1280\times 960$ but for depth, it is in $240\times180$. The ``Up-projected Sparse Depth'' means upsampling the raw depth map to the larger RGB image resolution, and this process will result in the sparsity of the depth map. See the ``Zoomed-in Region'' for better visualization. 
\label{Fig: huawei ToF sensor.}}
\end{figure*}

\begin{table*}
\centering
\begin{tabular}{c|lccc}
\toprule
\multicolumn{1}{l|}{}                                                                & Dataset & \multicolumn{1}{c}{Sparsity Pattern}                                              & Scenes                                                                        & Images \# \\ \midrule
\multirow{3}{*}{\begin{tabular}[c]{@{}c@{}}GeneralSparsity\\ Benchmark\end{tabular}} & NYUv2~\cite{silberman2012indoor}   & \begin{tabular}[c]{@{}c@{}}Sparse ToF / Unpaired FoV \\ / Short Range\end{tabular} & Indoor                                                                        & 654           \\
& ScanNet~\cite{dai2017scannet} & \begin{tabular}[c]{@{}c@{}}Sparse ToF / Unpaired FoV \\ / Short Range\end{tabular} & Indoor                                                                        & 700           \\
& DIODE~\cite{vasiljevic2019diode}   & \begin{tabular}[c]{@{}c@{}}Sparse ToF / Unpaired FoV \\ / Short Range\end{tabular} & \begin{tabular}[c]{@{}c@{}}Indoor\\ /Outdoor\end{tabular} & 771           \\ \midrule
\begin{tabular}[c]{@{}c@{}}NoisySparsity\\ Benchmark\end{tabular}                    & NYUv2   & \multicolumn{1}{l}{Holes, Noisy}                                                  & Indoor                                                                        & 4046          \\ \bottomrule
 \end{tabular}
\caption{Details of proposed two new benchmarks. `GeneralSparsity' aims to evaluate the model's robustness to different sparsity patterns across domains, while `NoisySparsity' evaluates the robustness to noises. \label{Tab: Two new benchmarks.}}
\end{table*}

\begin{figure*}[t]
\centering
\includegraphics[width=1\linewidth]{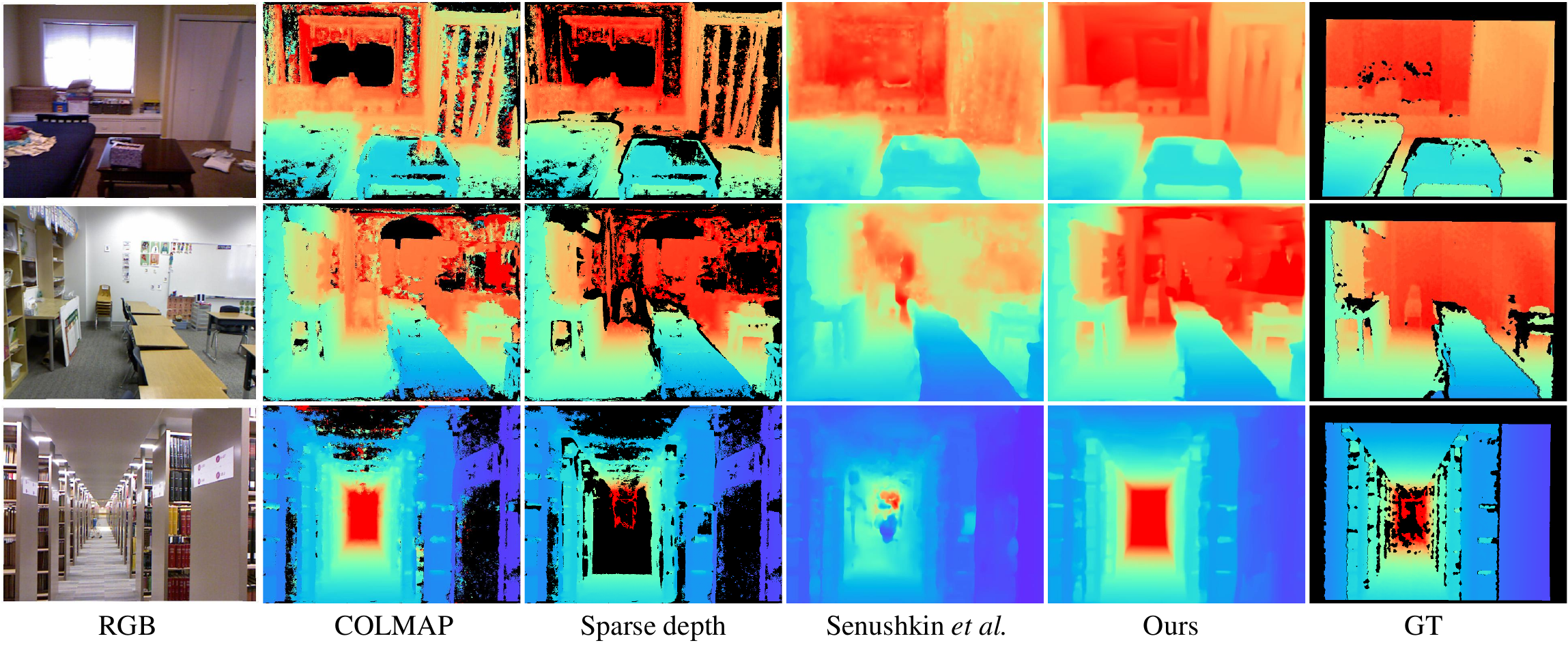}
\caption{Qualitative comparison for completing noisy sparse depth. The noisy sparse depths are obtained by masking COLMAP~\cite{schoenberger2016mvs} depths. Our completed results have fewer outliers and errors.\label{Fig: complete noisy depth visually.}}
\end{figure*}

\subsection{Benchmark Redesign}

Existing benchmarks such as NYU and Matterport3D only consider a specific type of sparsity, and thus they are not suitable for benchmarking a model's robustness and generalization ability across different task domains. 
Therefore, we propose two new benchmark designs based on typical types of depth sparsity and depth noise in real-world applications.
Critically, we note that the methods we use for generating sparse and noisy data are \emph{not} the same as those used for training our approach, this allows us to evaluate the generalization of our training sparsity patterns on these (more realistic) sparsity patterns.
Additionally, neither of these datasets was used when training our method.

The first one aims to evaluate the generalization ability of depth completion methods to various sparsity situations, termed \emph{GeneralSparsity}.
We collect 3 datasets (NYU~\cite{silberman2012indoor}, ScanNet~\cite{dai2017scannet}, as well as  DIODE~\cite{vasiljevic2019diode}) and propose 3 sparse depth patterns to simulate commodity-level RGBD sensors: 1) \emph{Unpaired FOV}. Some RGBD sensors have different fields of view between the RGB camera and the depth sensor, such as KinectV2 (RGB: $84.1^\circ \times 53.8^\circ$
; Depth: $70.6^\circ \times 60^\circ$
)~\cite{wasenmuller2016comparison}. 
We propose to mask a $25\%$ region along the $4$ borders of the ground truth depth to simulate this situation. 
2) \emph{Sparse ToF}.  Time-of-light sensors (ToF) embedded in mobile devices have much lower resolution than that of an RGB camera. 
For example, the RGB sensor and the depth sensor on a Huawei phone have the same field of view but the resolutions for the RGB camera and the ToF sensor are $1280\times 960$ and $240\times180$ respectively. 
We up-project the raw depth to the RGB size and obtain the sparse depth shown in Fig.~\ref{Fig: huawei ToF sensor.}. 
In our benchmark, we downsample the ground truth depth map to a low resolution, up-project to the original size and mask the distant regions to obtain the sparse depth. 3) \emph{Short Range}. 
To simulate the short measuring range of depth sensors (such as RealSense), we mask the $50\%$ most distant regions of ground truth to obtain the incomplete depth. 

Furthermore, we set up another benchmark to evaluate the robustness to noisy input, termed \emph{NoisySparsity}. 
We sample 16 videos with over 4000 images from NYU. 
To simulate the noise pattern in a real scene reconstruction application, COLMAP~\cite{schoenberger2016mvs} is used to reconstruct the scene, and we mask the noisiest regions based on the inconsistency between ground truth depth and the COLMAP depth. 
Note that the masked incomplete depths are still very noisy. 
Some examples are shown in Fig.~\ref{Fig: complete noisy depth visually.}. More details of these two benchmarks are shown in Table~\ref{Tab: Two new benchmarks.}.

\section{Experiments}
In this section, we conduct several experiments to demonstrate the effectiveness of our approach.

\noindent\textbf{Implementation Details.}
We employ the network proposed by Seichter~\etal~\cite{esanet2020} with a pretrained ResNet-34~\cite{he2016deep} backbone in our experiments.
We use SGD for optimization with an initial learning rate of $0.02$ for all layers.
The learning rate is decayed every $40000$ iterations with the ratio $0.1$, and we used 24 samples per batch. We sampled $12000$ images from three datasets respectively for training, i.e. Taskonomy~\cite{zamir2018taskonomy}, DIML~\cite{kim2018deep}, and TartanAir~\cite{wang2020tartanair}. 
During training, images are randomly flipped horizontally and resized to $448\times448$. 

\noindent\textbf{Evaluation Metrics.} Following previous methods~\cite{park2020non, cheng2020cspn++}, we employ multiple metrics to evaluate the results, including absolute relative error (AbsRel), mean absolute error (MAE), root mean squared error (RMSE), and percentage of pixels satisfying $\delta_{\tau} =  max\left ( \frac{d_{pred}}{d_{gt}}, \frac{d_{gt}}{d_{pred}} \right )< 1.25^{\tau} $. For runtime analysis, we only record the runtime of the model forward process, and regard the average of 3 times of 1000 forward process on a single Nvidia RTX 3090 Ti as `Time'.

\begin{table*}[!ht]
\centering
\setlength{\tabcolsep}{12pt}
\resizebox{\linewidth}{!}{
\begin{tabular}{l|llllll}
\toprule
Methods                & RMSE$\downarrow$ & AbsRel$\downarrow$ & $\delta_{1}\uparrow$     & $\delta_{2}\uparrow$      & $\delta_{3}\uparrow$ & Time$\downarrow$       \\ \midrule
S2D~\cite{ma2018sparse}        & $0.230$   & $0.044$   & $97.1$ & $99.4$ & $99.8$ & -   \\
S2D+SPN~\cite{liu2017learning} & $0.172$   & $0.031$   & $98.3$ & $99.7$ & $99.9$ & - \\
DepthCoeff~\cite{imran2019depth}   & $0.118$   & $0.013$   & $99.4$ & $99.9$ & - & -     \\
CSPN~\cite{cheng2018depth}         & $0.117$   & $0.016$   & $99.2$ & $99.9$ & $100.0$ & 43.45ms \\
CSPN++~\cite{cheng2020cspn++}       & $0.116$   & -       & -    & -    & -  &  -  \\
DeepLiDAR~\cite{qiu2019deeplidar}    & $0.115$   & $0.022$   & $99.3$ & $99.9$ & $100.0$ & 207.5ms \\
DepthNormal~\cite{xu2019depth} & $0.112$   & $0.018$   & $99.5$ & $99.9$ & $100.0$ & - \\
NLSPN~\cite{park2020non}                 & $0.092$   & $\textbf{0.012}$   & $99.6$ & $99.9$ & $100.0$& \textbf{16.3ms} \\ 
Lee~\etal~\cite{lee2021depth} &$0.104$ &$0.014$ &$99.4$ &$99.9$ &$100.0$ & - \\
Huynh~\etal~\cite{huynh2021boosting}& $\textbf{0.090}$   & $0.014$   & $\textbf{99.6}$ & $\textbf{99.9}$ & $\textbf{100.0}$ & - \\ \midrule
MiDaS~\cite{Ranftl2020}  &$0.513$  &$0.110$ &$88.6$ &$98.1$ &$99.6$  &  17.1ms    \\
Yin~\etal~\cite{Wei2021CVPR} &$0.402$  &$0.090$ &$91.3$ &$98.0$ &$99.5$  &  26.0ms    \\ \midrule
Ours-baseline &$0.210$  &$0.036$ &$98.4$ &$99.6$ &$99.9$  & 23.3ms     \\
Ours-W MiDaS~\cite{Ranftl2020}   &$0.199$  &$0.024$ &$98.6$ &$99.6$ &$99.9$  & 23.3ms+17.1ms      \\
Ours-W Yin~\etal~\cite{Wei2021CVPR}   &$0.183$  &$0.022$ &$98.7$ &$99.7$ &$99.9$  & 23.3ms+26.0ms       \\ \bottomrule
\end{tabular}
}
\caption{Depth completion results on the NYU dataset. Following ~\cite{cheng2018depth, park2020non}, we uniformly sample 500 points from ground truth as the sparse depth. Our method is \emph{not trained on NYU} but is comparable with state-of-the-art methods that \emph{are} trained on NYU. MiDaS and Yin~\etal are monocular depth estimation methods that are used as input to our method. ``23.3ms+17.1ms'' means depth completion network requires 23.3ms and the generation of guidance map spends 17.1ms.}
\label{Tab: DepComplete Cmp with SOTA on NYU}
\end{table*}

\begin{table*}[t]
\centering
\setlength{\tabcolsep}{8pt}
\begin{tabular}{l|lllllll}
\toprule
Methods           & RMSE$\downarrow$  & MAE$\downarrow$   & $\delta_{1.05}\uparrow$ & $\delta_{1.1}\uparrow$  & $\delta_{1}\uparrow$  & $\delta_{2}\uparrow$ & $\delta_{3}\uparrow$  \\ \midrule
Huang~\etal~\cite{huang2019indoor} & $1.092$ & $0.342$ & $66.1$ & $75.0$ & $85.0$ & $91.1$ & $93.6$ \\
Zhang~\etal~\cite{zhang2018deepdepth}& $1.316$ & $0.461$ & $65.7$ & $70.8$ & $78.1$ & $85.1$ & $88.8$    \\
Senushkin~\etal~\cite{senushkin2020decoder}& $\textbf{1.028}$ & $\textbf{0.299}$ & $\textbf{71.9}$ & $\textbf{80.5}$ & $\textbf{89.0}$ & $\textbf{93.2}$  & $95.0$  \\ \midrule
Yin~\etal~\cite{Wei2021CVPR} & $2.06$  & $1.13$ & $17.9$ & $29.8$ & $50.7$   & $72.3$ & $83.4$      \\
MiDaS~\cite{Ranftl2020}  & $3.45$  & $2.01$ & $13.2$ & $21.8$ & $37.5$   & $54.8$ & $66.4$      \\\midrule
Ours-baseline & $2.35$ & $0.574$ & $68.9$ & $78.6$ & $86.1$ & $91.5$ & $96.0$                             \\
Ours-W MiDaS \cite{Ranftl2020}   & $1.49$ & $0.448$ & $67.8$ & $76.3$ & $85.0$ & $91.0$ & $94.5$                           \\
Ours-W Yin \etal \cite{Wei2021CVPR}   & $\underline{1.03}$ & $\underline{0.320}$ & $\underline{71.2}$ & $\underline{79.0}$ & $\underline{87.1}$ & $\underline{93.1}$ & $\textbf{96.0}$                           \\ \bottomrule
\end{tabular}
\caption{Depth completion results on the Matterport3D dataset. Our method is \emph{not trained on Matterport3D} but is comparable with state-of-the-art methods that are trained on Matterport3D. RMSE and MAE are given in meters. MiDaS and Yin~\etal are monocular depth estimation methods that are used as input to our method.}
\label{Tab: DepComplete Cmp with SOTA on Matterport3D}
\end{table*}

\subsection{Depth Completion Evaluation}
In this experiment, we first compare our method with previous methods on the old NYU and Matterport3D benchmarks to show that our method can achieve performance comparable to previous methods, although it has not been trained on either of these datasets.
Second, we benchmark on our new sparse depth completion benchmarks.
Finally, we test our method on the DualPixel dataset~\cite{GargDualPixelsICCV2019} captured by a smartphone to further show the generalization of our method to mobile sensors. 

\begin{figure*}[t]
\centering
\includegraphics[width=1\linewidth]{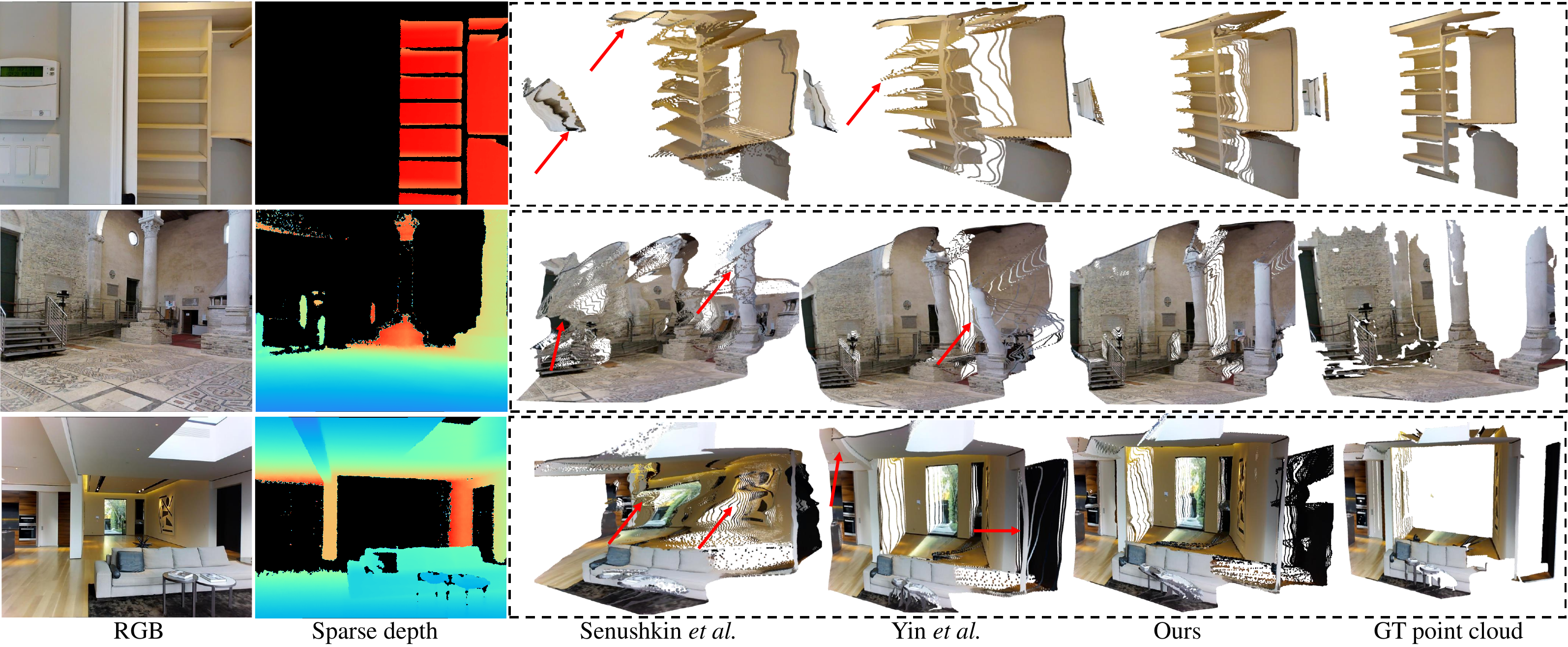}
\caption{Qualitative comparison of depth and reconstructed 3D shape. Our completed metric depth has finer details and the reconstructed 3D shape is more accurate than Senushkin~\etal~\cite{senushkin2020decoder} and Yin~\etal~\cite{Wei2021CVPR}. Note that Yin~\etal cannot predict metric depth.}
\label{Fig: cmp of depth and pcd on matterport3d.}
\end{figure*}

\begin{table*}[!t]
\centering
\scriptsize
\begin{tabular}{c|cc|cccc} \toprule
  \multicolumn{3}{c|}{Methods} & NLSPN~\cite{park2020non} & Senushkin~\etal~\cite{senushkin2020decoder} & Ours-baseline & Ours-W Guidance \\ \midrule
  \multirow{3}{*}{NYU} & \multirow{3}{*}{AbsRel$\downarrow$} & Unp. FOV & $0.150$ & $0.224$ & $0.046$ & $\textbf{0.031}$ \\
  & & Sparse ToF & $0.190$ & $0.615$ & $0.018$ & $\textbf{0.013}$ \\
  & & Short Range & $0.114$ & $0.093$ & $0.041$ & $\textbf{0.030}$ \\ \midrule
  \multirow{3}{*}{ScanNet} & \multirow{3}{*}{AbsRel$\downarrow$} & Unp. FOV & $0.716$ & $0.255$ & $0.049$ & $\textbf{0.028}$ \\
  & & Sparse ToF & $1.413$ & $0.793$ & $0.022$ & $\textbf{0.014}$ \\
  & & Short Range & $0.202$ & $0.166$ & $0.047$ & $\textbf{0.037}$ \\ \midrule
   \multirow{3}{*}{DIODE} & \multirow{3}{*}{AbsRel$\downarrow$} & Unp. FOV & $6.684$ & $0.687$ & $0.150$ & $\textbf{0.139}$ \\
  & & Sparse ToF & $11.370$ & $6.120$ & $0.143$ & $\textbf{0.111}$ \\
  & & Short Range & $1.005$ & $0.623$ & $0.144$ & $\textbf{0.137}$ \\ \bottomrule
\end{tabular}
\caption{Comparison of state-of-the-art methods on our proposed GeneralSparsity benchmark. This benchmark mainly analyzes the robustness of depth completion methods to different sparsity types. Although these testing datasets and sparsity types are unseen to our methods during training, we can achieve better performance than current methods. 
\label{Tab: Generalization cmp.}}
\end{table*}

\noindent\textbf{Existing Benchmarks.}
Here, we compare to current state-of-the-art methods on NYU~\cite{silberman2012indoor} and Matterport3D~\cite{Matterport3D}.
Although these two benchmarks have different sparse depth types, we use a single model for the evaluation on both of them.
We include a baseline method (Ours-baseline) where the model does not make use of depth guidance, that is, it directly predicts the complete depth from RGB and sparse depth. 
Results are shown in Table~\ref{Tab: DepComplete Cmp with SOTA on NYU} and Table~\ref{Tab: DepComplete Cmp with SOTA on Matterport3D}, which indicate that our method achieves performance comparable to current methods, despite not being trained on these datasets.

\begin{figure*}
\centering
\includegraphics[width=1\linewidth]{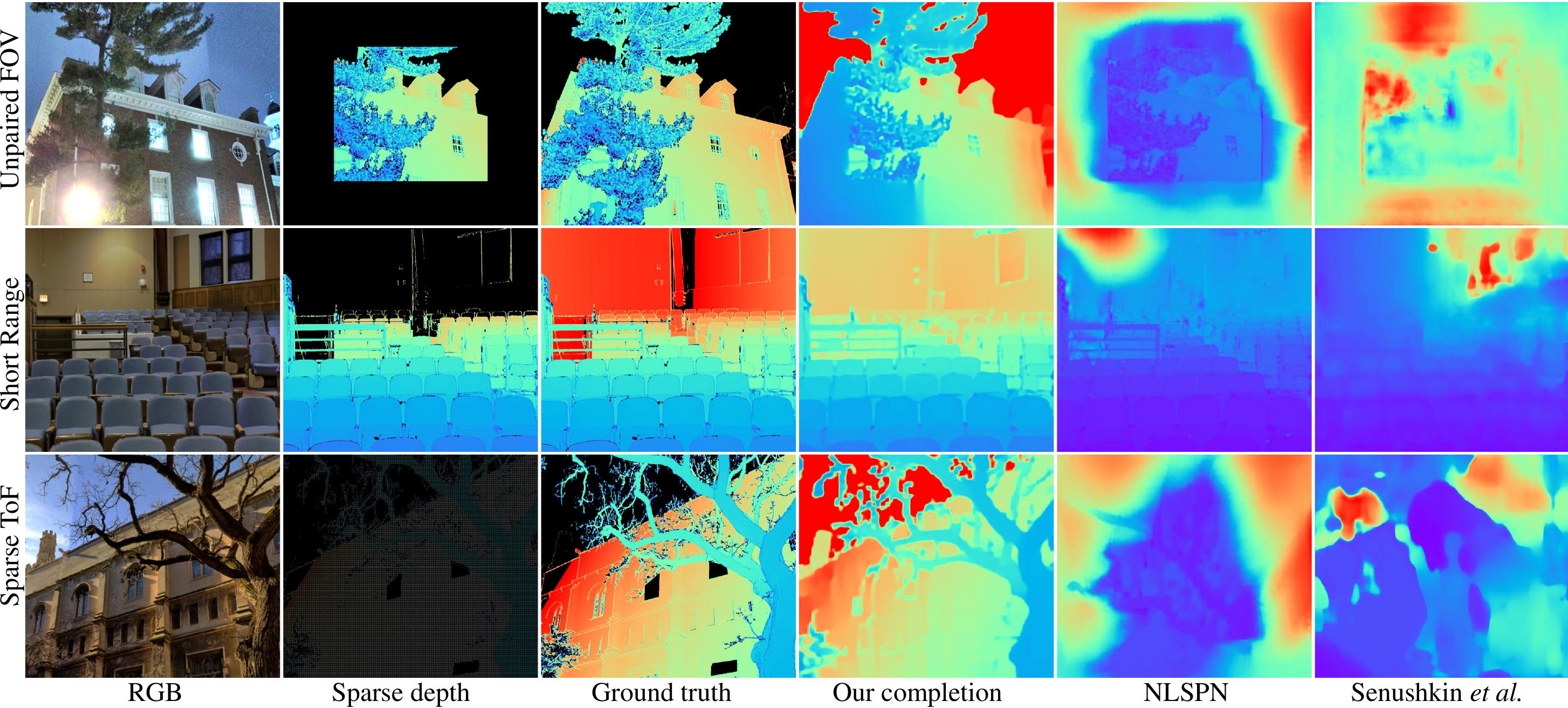}
\caption{Qualitative completion results on the DIODE~\cite{vasiljevic2019diode} dataset. Note that none of the methods are trained on this dataset. We compare our method with Senushkin~\etal~\cite{senushkin2020decoder} and NLSPN~\cite{park2020non} using $3$ different unseen sparsity patterns.
\label{Fig: diff sparsity cmp}}
\end{figure*}

Following ~\cite{cheng2018depth, park2020non}, we uniformly sample 500 points from the ground truth as the sparse depth on NYU.
We can see that our method achieves performance on par with previous methods, and better accuracy than the baseline.
Compared with our baseline (Ours-baseline), using a pretrained guidance model further improves performance. 
When compared to Yin~\etal~\cite{Wei2021CVPR} (which provides our guidance input), we see that even with only 500 sparse points, the performance is improved significantly.
Furthermore, we observe that our method is robust with respect to different guides, see `Ours-W MiDaS' (using the MiDaS depth as the input during testing) and `Ours-W Yin~\etal~\cite{Wei2021CVPR}'.
As depths from Yin~\etal~\cite{Wei2021CVPR} are employed for the guidance map during training, taking their depth can achieve better results in testing.

Existing methods were trained on small-scale data. To the best of our knowledge, our work is the first to prove the feasibility of our robust monocular depth completion paradigm, and it is one of our contributions. The performance difference between the compared methods and ``Ours-baseline'' mainly comes from the effectiveness of the training paradigm on diverse datasets, and the performance difference between ``Ours-baseline'' and ``Ours-W Guidance'' mainly comes from our guidance map pipeline. As a result, our comparisons in Table 5 can show the effectiveness of both the robust monocular depth completion paradigm and the guidance-map pipeline. Also, the performance of methods with the guidance map can outperform those of depth maps outputted by the depth prediction network MiDaS and Yin \etal by a large margin, and it ensures the impact and necessity of our guidance map pipeline.

It is worth noticing that our analysis in Fig.~\ref{Fig: Robustness cmp.} has shown the lack of robustness of existing depth completion methods, and our reliability in practical applications. For Table~\ref{Tab: DepComplete Cmp with SOTA on NYU} and Table~\ref{Tab: DepComplete Cmp with SOTA on Matterport3D}, the upper part of the methods are not only trained on the training split of test datasets, but also lack robustness to outliers and sample point numbers. Even though we report their best performance on suitable point numbers without any noise of ground truth, we can still achieve comparable performance without training on the domain of test datasets. If we offer noisy ground-truth depth, our methods can outperform these methods by a large margin, according to Table~\ref{Tab: Generalization cmp.} and Table~\ref{Tab: Noisy cmp.}.

Moreover, the qualitative comparison on Matterport3D is illustrated in Fig.~\ref{Fig: cmp of depth and pcd on matterport3d.}.
Although Senushkin~\etal~\cite{senushkin2020decoder} can achieve better accuracy than ours (note that it was trained on Matterport3D) 
we find that our reconstructed scene structure is better.

\noindent\textbf{Proposed New Benchmarks.}
We now compare to current state-of-the-art methods on the proposed GeneralSparsity and NoisySparsity datasets.

\begin{figure*}[t]
\centering
\includegraphics[width=\linewidth]{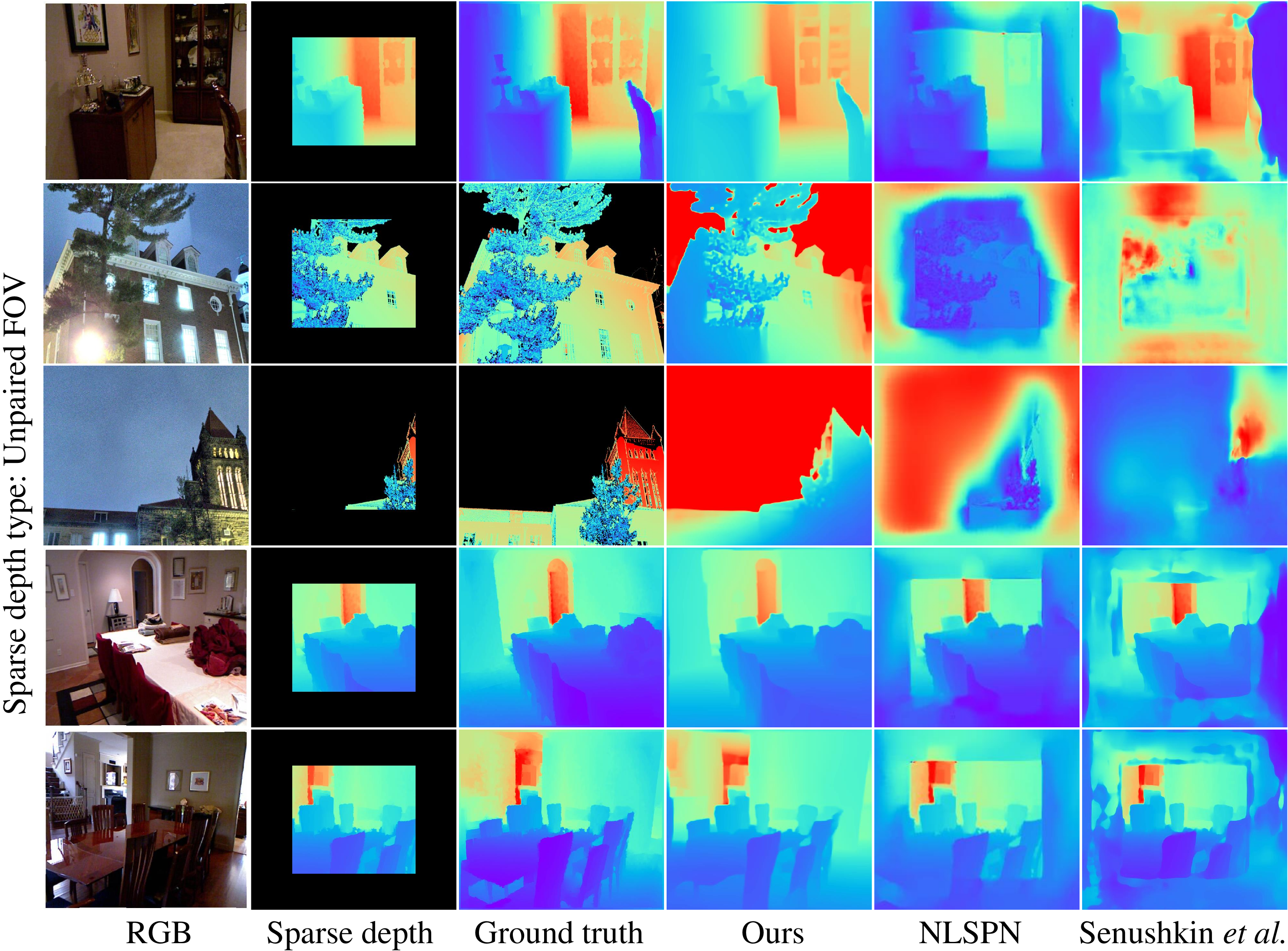}
\caption{The comparison on completing `Unpaired FOV' sparse depth. We create this type by masking $25\%$ region around $4$ borders of the ground truth depth for evluation. We can see that our method is more robust on this type. \label{Fig: unpaired FOV comparison}}
\vspace{-1.5em}
\end{figure*}

\begin{figure*}[t]
\centering
\includegraphics[width=\linewidth]{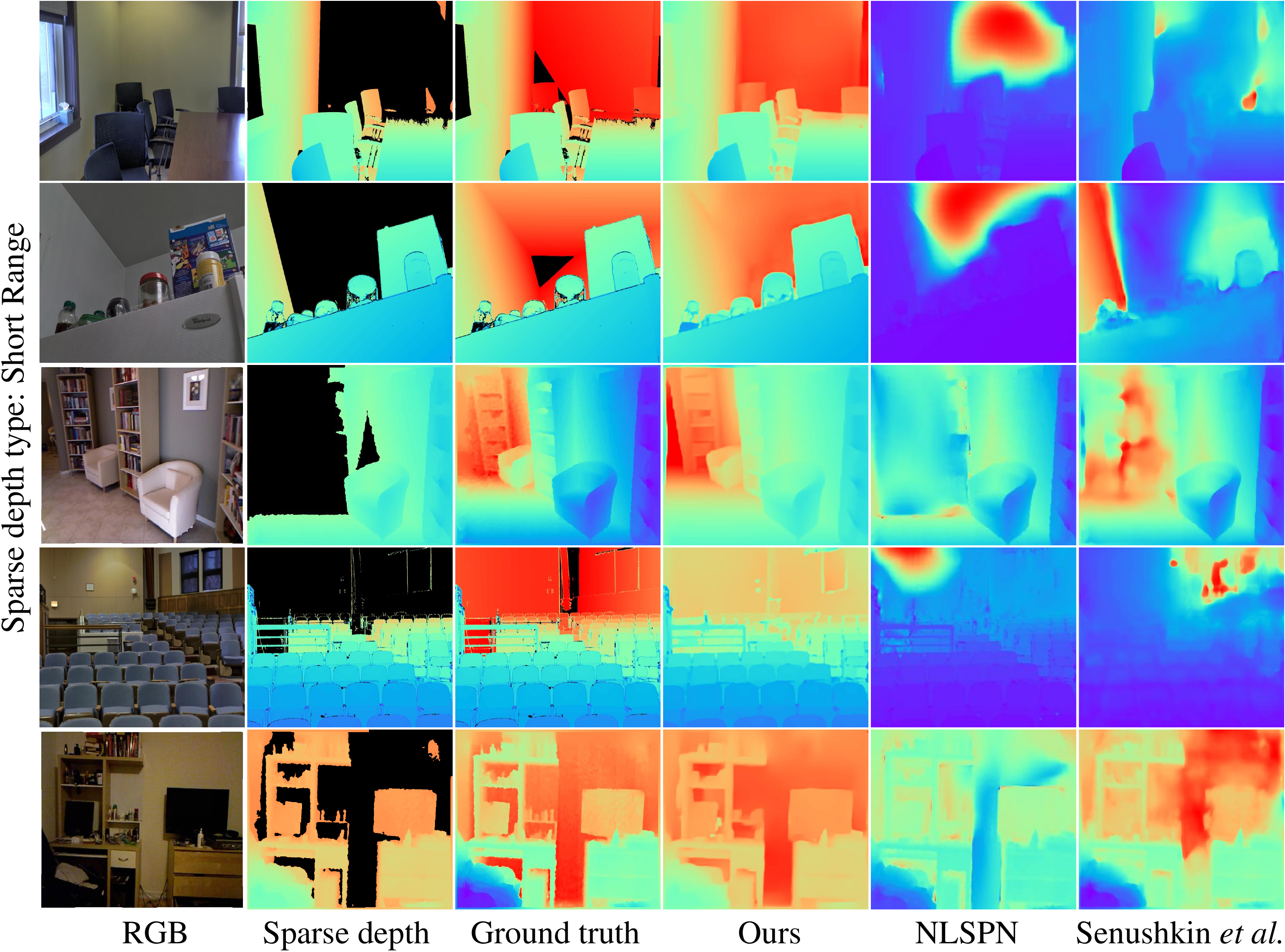}
\caption{The comparison on completing `Short Range' sparse depth. We create this type by masking the most $50\%$ distant regions. We can see that our method is more robust on this type. \label{Fig: no distant}}
\vspace{-1.5em}
\end{figure*}

\begin{figure*}[t]
\centering
\includegraphics[width=\linewidth]{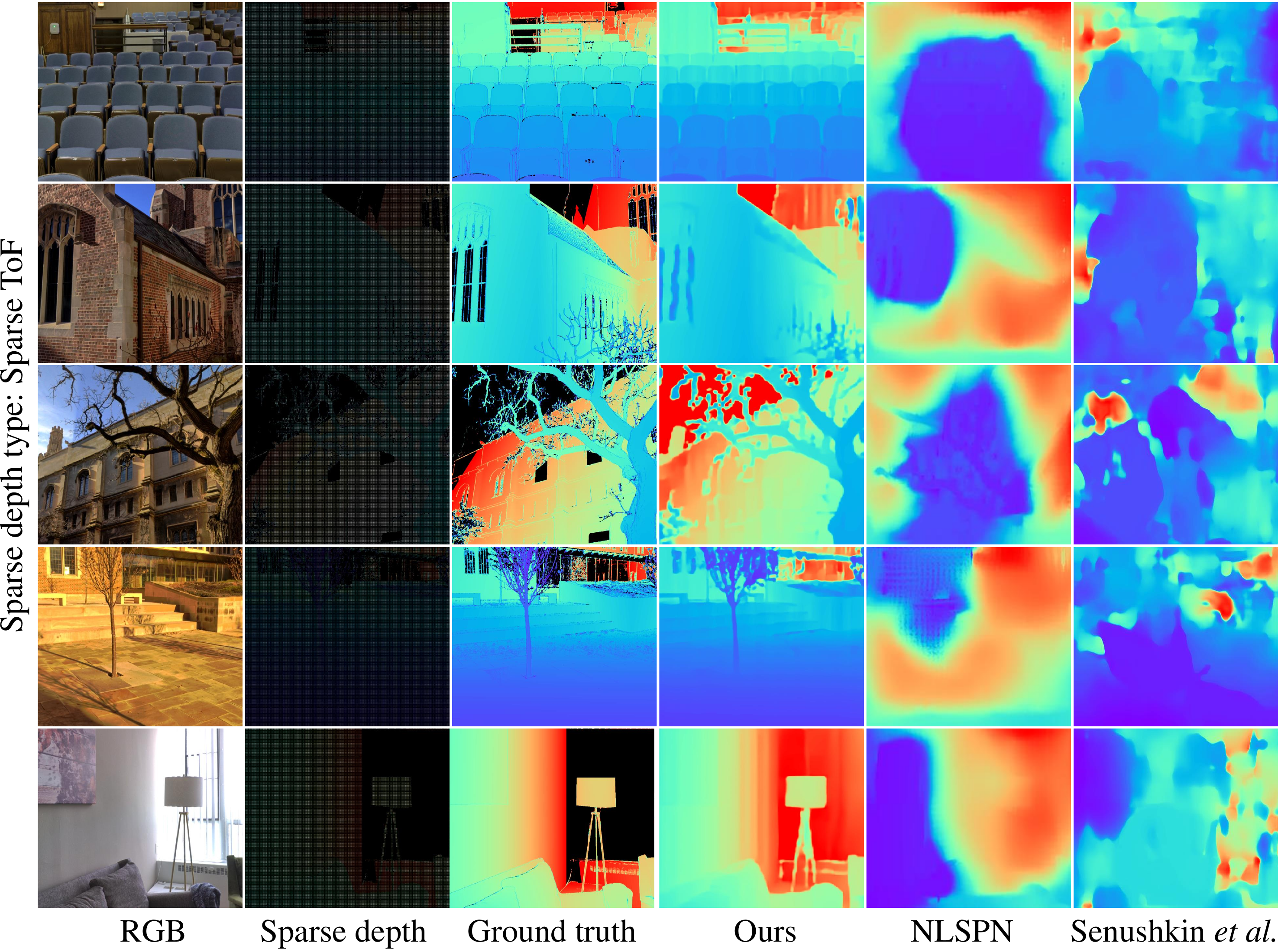}
\caption{The comparison on completing `Sparse ToF' sparse depth. We create this type by sampling the depth from the ground truth every $3$ pixels horizontally and vertically. We can see that our method can acheive better results.  \label{Fig: sparse tof comparison}}
\vspace{-1.5em}
\end{figure*}

We compare our approach to NLSPN~\cite{park2020non} and Senushkin~\etal~\cite{senushkin2020decoder} since they have achieved the most promising results on current benchmarks.
Note that the NLSPN method is trained on NYU, while Senushkin~\etal~\cite{senushkin2020decoder} is trained on Matterport3D.
In contrast, our method is not trained on either of these datasets. Furthermore, the specific sparsity patterns used for evaluation are not utilized in our training.
Results are summarized in Table~\ref{Tab: Generalization cmp.}. 
We find that NLSPN~\cite{park2020non} and Senushkin~\etal~\cite{senushkin2020decoder} do not generalize well to different types of sparse depth, or datasets, which is reasonable as they are trained on one dataset and one sparsity type only.
In contrast, our method can achieve comparable performance on both datasets. 
Furthermore, compared to our baseline method, we see that using the guidance map can consistently improve the performance over all datasets and sparse depth types.
The quantitative comparison of completing $3$ different patterns on the DIODE dataset is demonstrated in Fig.~\ref{Fig: diff sparsity cmp}.
More examples on our proposed GeneralSparsity benchmark are shown in Fig.~\ref{Fig: unpaired FOV comparison}, \ref{Fig: no distant}, and \ref{Fig: sparse tof comparison}.

\begin{figure}
\centering
\includegraphics[width=\linewidth]{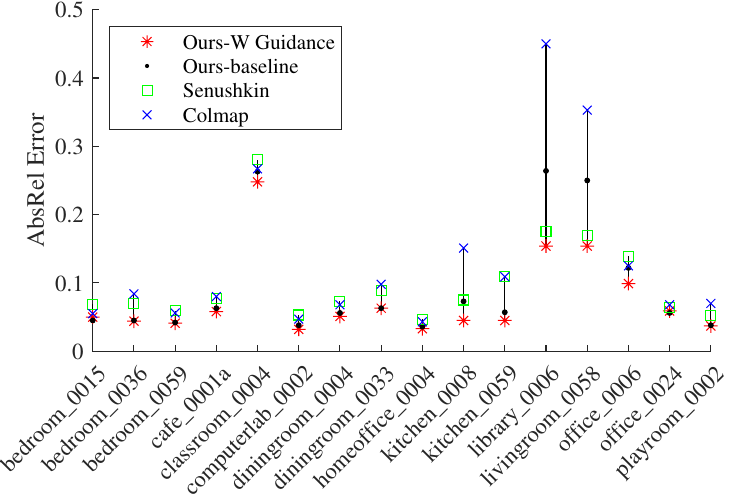}
\caption{Completion results on $16$ scenes sampled from NYU~\cite{silberman2012indoor}. The input depth is noisy, which is generated using COLMAP~\cite{schoenberger2016mvs}. \label{Fig: complete noisy depth.}}
\end{figure}

\begin{figure*}[t]
\centering
\includegraphics[width=\linewidth]{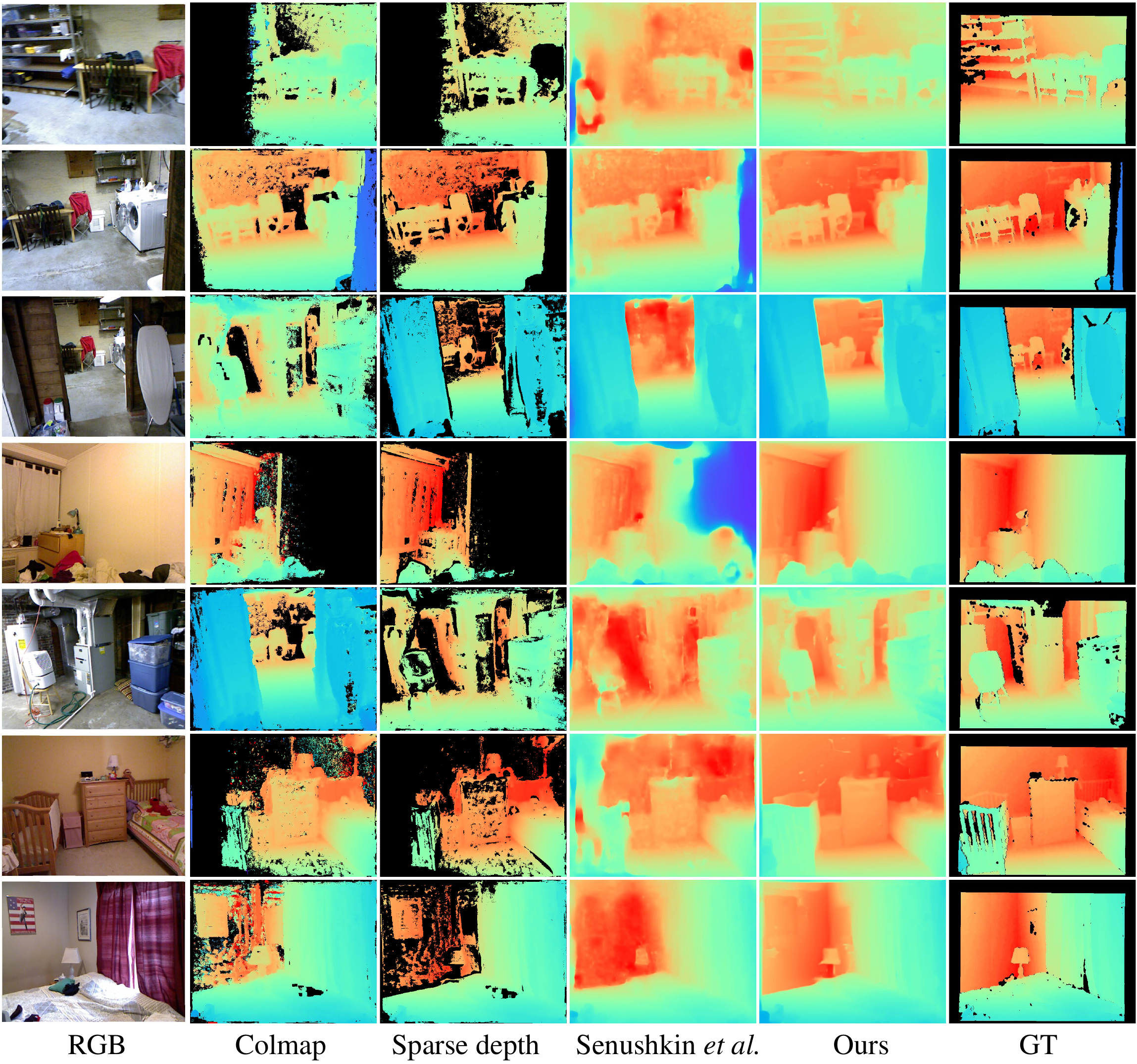}
\caption{Complete noisy depth maps on sampled $16$ NYU scenes. We use the colmap~\cite{schoenberger2016mvs} to create the noisy sparse depth for completion. Our method is consistently better than other methods on all scenes. \label{Fig: complete noisy depths 1}}
\vspace{-1.5em}
\end{figure*}

\begin{figure*}[t]
\centering
\includegraphics[width=\linewidth]{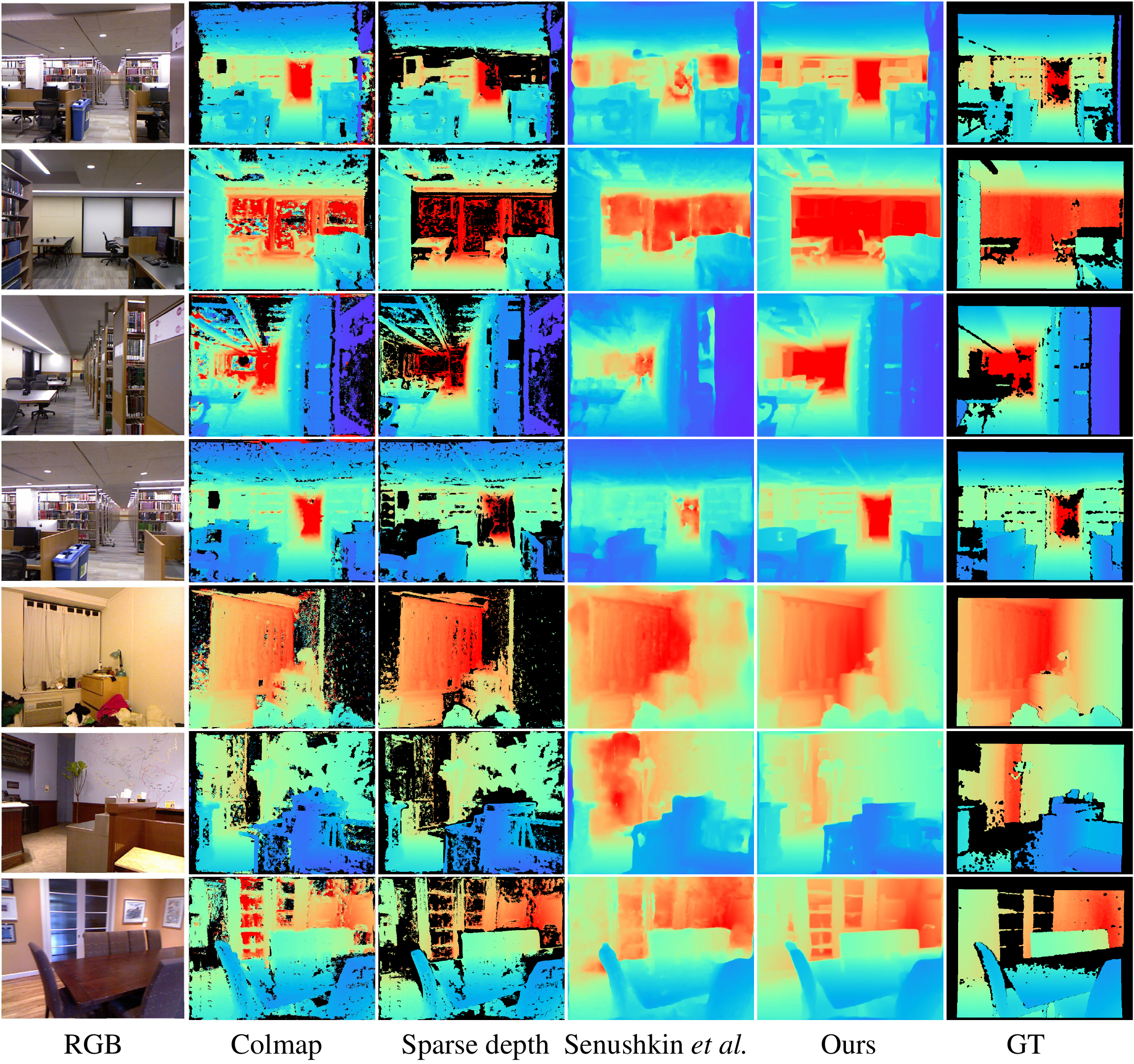}
\caption{Complete noisy depth maps on sampled $16$ NYU scenes. We use the colmap~\cite{schoenberger2016mvs} to create the noisy sparse depth for completion. Our method is consistently better than other methods on all scenes. \label{Fig: complete noisy depths 2}}
\vspace{-1.5em}
\end{figure*}

On the NoisySparsity benchmark, we evaluate robustness to noisy sparse inputs.
In Table~\ref{Tab: Noisy cmp.}, we observe that our method can achieve better accuracy than existing methods.
Compared with our baseline approach, using the guidance map further boosts performance. 
Qualitative comparisons are shown in Fig.~\ref{Fig: complete noisy depth visually.}, we can see that our completed depths have much fewer outliers and noise (see the wall).
More quantitative comparisons can be found in Fig.~\ref{Fig: complete noisy depth.}, \ref{Fig: complete noisy depths 1} and \ref{Fig: complete noisy depths 2}. Compared with the baseline (no guidance map), our approach is more robust to noise (see `library\_0006' and `living\_room\_0058').

\begin{table*}[!ht]
\centering
\begin{tabular}{l|cc|cc}
\toprule
Metrics
           &\begin{tabular}[c]{@{}c@{}}NLSPN\\ \cite{park2020non}\end{tabular}  & \begin{tabular}[c]{@{}c@{}}Senushkin\etal\\ ~\cite{senushkin2020decoder}\end{tabular} & \begin{tabular}[c]{@{}c@{}}Ours\\ (baseline)\end{tabular} & \begin{tabular}[c]{@{}c@{}}Ours\\ (W Guidance)\end{tabular} \\ \midrule
AbsRel (\%)$\downarrow$     & 24.9 & 18.26     & 6.41                                                      & \textbf{5.25}                                                        \\
$\delta_1$ (\%) $\uparrow$ & 54.7  & 70.3      & 93.6                                                      & \textbf{95.2}                                                        \\ \toprule
\end{tabular}
\caption{Comparison on the proposed NoisySparsity benchmark, showing robustness to noisy input.
\label{Tab: Noisy cmp.}}
\end{table*}

\begin{figure*}[t]
\centering
\includegraphics[width=1\linewidth]{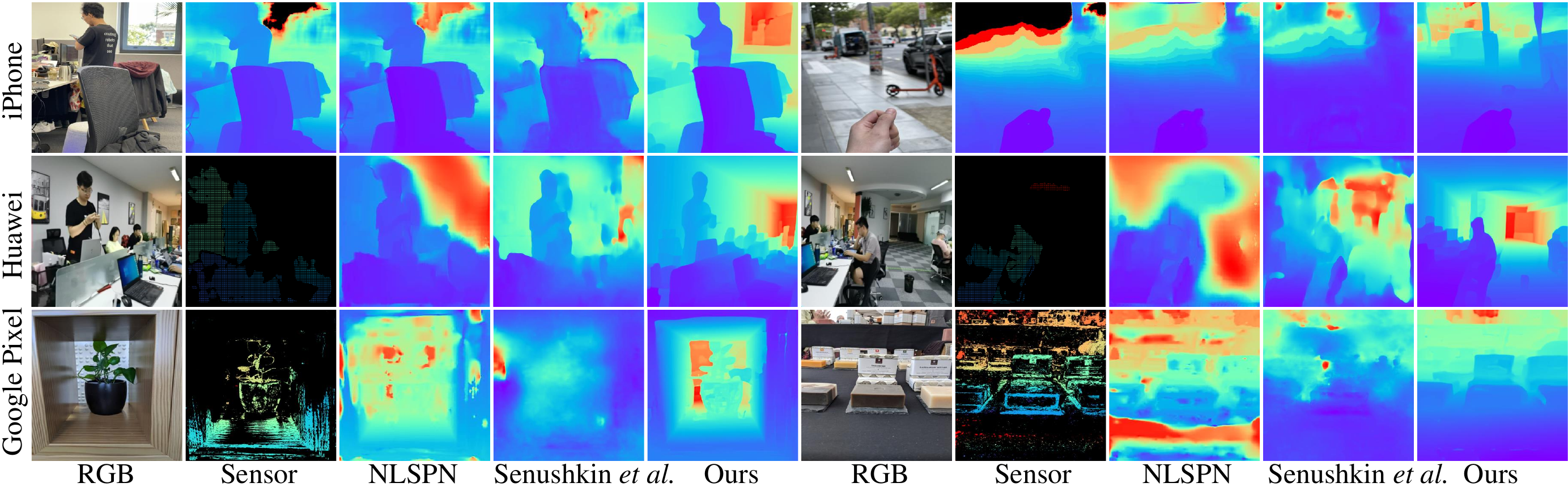}
\caption{Completion of the phone-captured depths. Our method is more robust to different depth sensors than previous methods.}
\label{Fig: sensor complete cmp.}
\end{figure*}

\begin{table*}[t]
\centering
\begin{tabular}{l|ccc}
\toprule
 \multirow{3}{*}{Methods}
& \multicolumn{3}{c}{AbsRel $\downarrow$} \\
             & \begin{tabular}[c]{@{}c@{}}NYU\\ (Sparse ToF)\end{tabular} & \begin{tabular}[c]{@{}c@{}}ScanNet\\ (Unpaired FOV)\end{tabular} &  \begin{tabular}[c]{@{}c@{}}Matterport3D\\ (Original pattern)\end{tabular} \\ \midrule
W/o Features &$0.036$   &$0.047$  &$0.122$ \\
W/o Uniform  &$0.021$ &$0.044$   & $0.121$ \\
W/o Holes    &$0.012$ &$0.12$  &$0.165$ \\
Ours-Full    &$0.013$   &$0.028$ &$0.120$  \\
\toprule
\end{tabular}
\caption{Ablation of different training sparsity patterns. The model is trained on the simulated patterns except the one specified by `W/o` and evaluated on zero-shot datasets.    \label{Table: different sparse patterns.}}
\end{table*}

\noindent\textbf{Completing Mobile Phone Sensor Depth.}
Many smartphones have been equipped with cheap 3D sensors.
To evaluate the robustness of our completion method with respect to noisy phone-captured depths, we perform a quantitative comparison on a Google Pixel 3  captured RGBD dataset (DualPixel)~\cite{GargDualPixelsICCV2019}.
We apply the provided confidence map to sample the most confident 36\% regions as the sparse depths and evaluate the depth on the most confident 80\% regions.
On this dataset, our method achieves \textbf{1.6\%} AbsRel ($\downarrow$), while NLSPN~\cite{park2020non} and Shenshkin~\cite{senushkin2020decoder} only achieve \textit{4.1\%} and \textit{24.86\%} respectively.

\begin{figure*}[t]
\centering
\includegraphics[width=\linewidth]{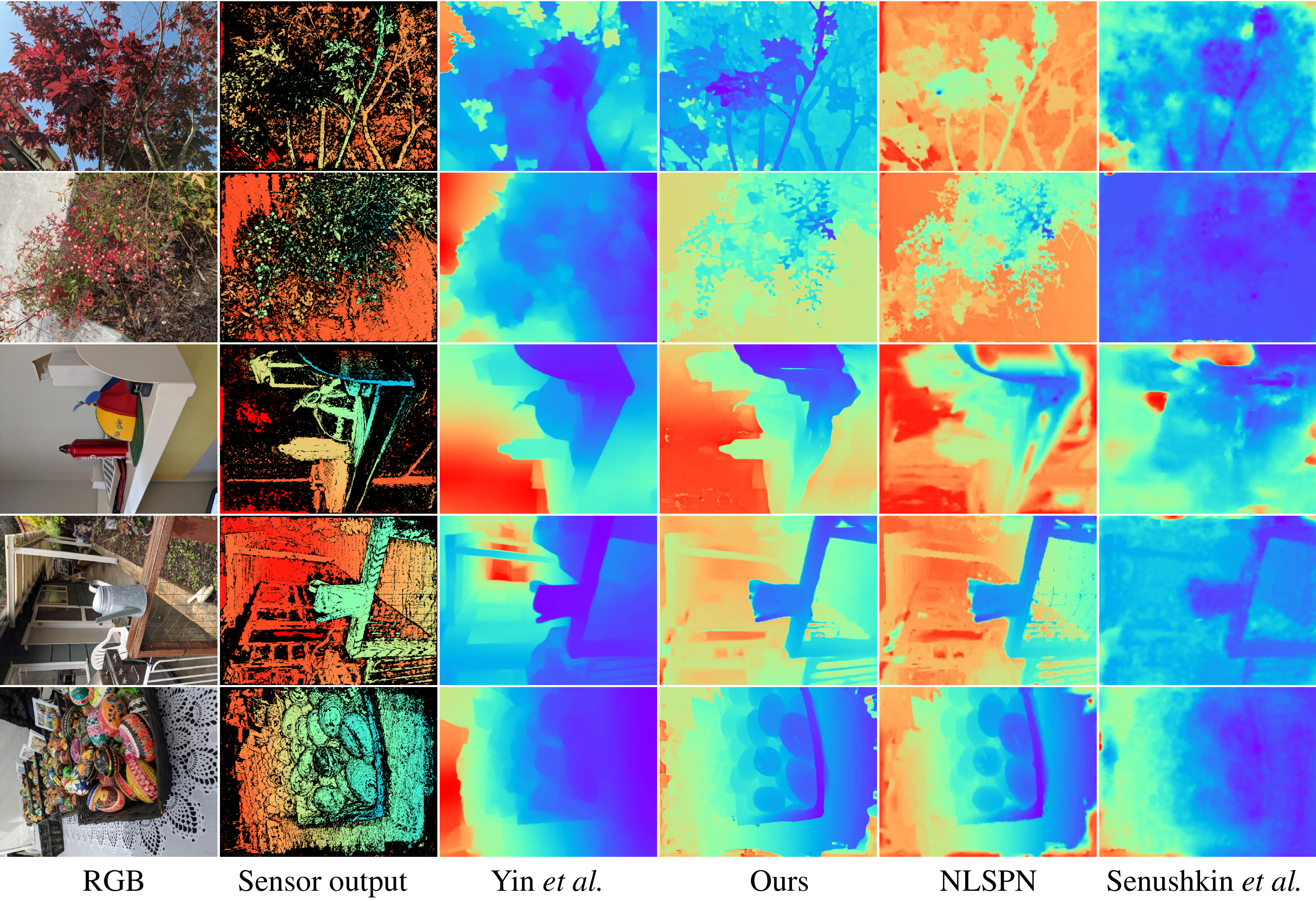}
\caption{The comparison of completing Pixel phone captured depths. We can see that our method can recover better results than other methods. \label{Fig: Pixel depth}}
\vspace{-1.5em}
\end{figure*}

\begin{figure*}[t]
\centering
\includegraphics[width=\linewidth]{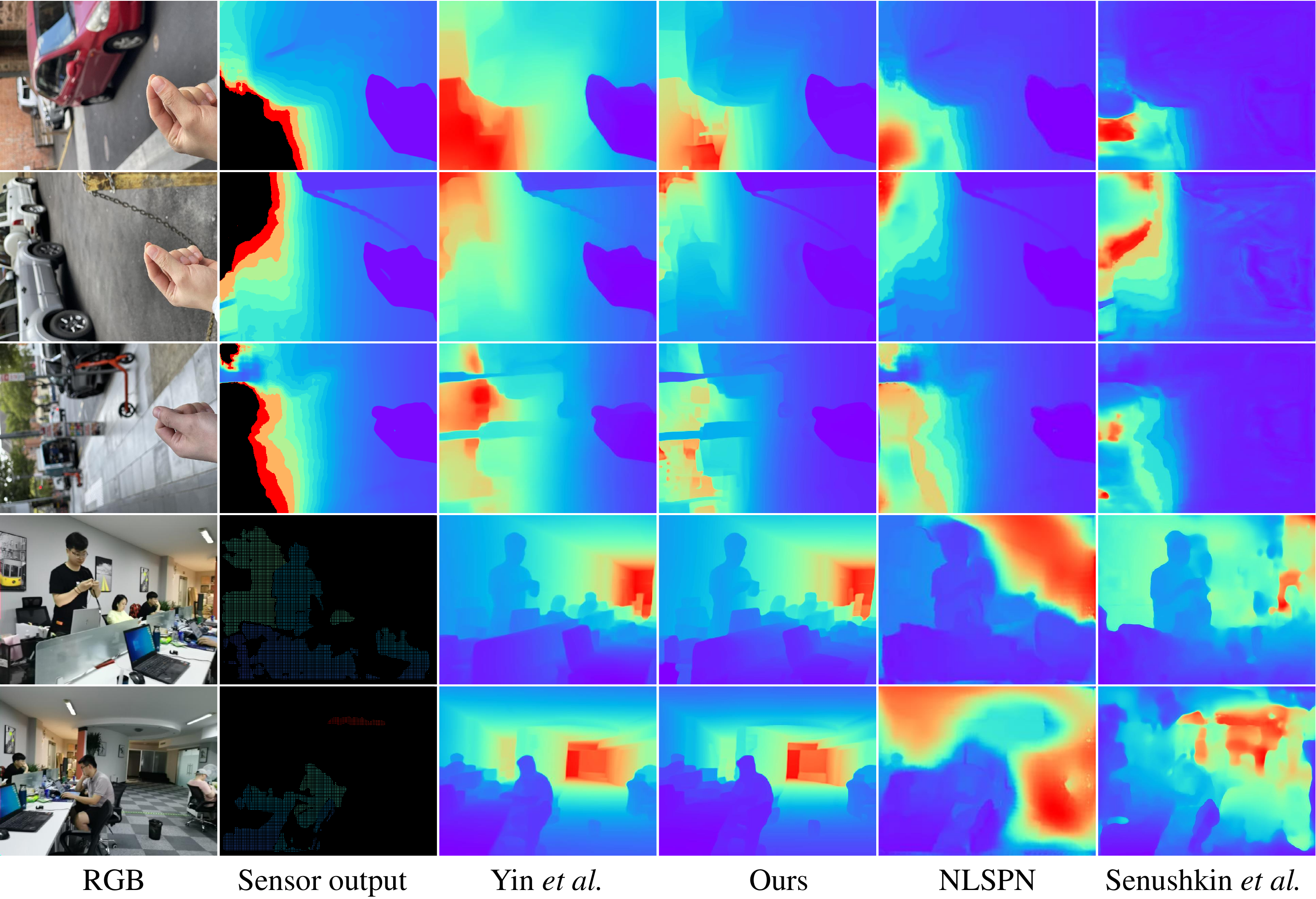}
\caption{The comparison of completing iPhone and Huawei phone captured depths. The first 3 rows examples are captured by iPhone, while the others are obtained by Huawei.  We can see that our method can recover better results than other methods. \label{Fig: iPhone and Huawei}}
\vspace{-1.5em}
\end{figure*}

Furthermore, we also qualitatively test on $2$ more phones, i.e. iPhone 12, and Huawei Mate 30.
In each case, the depths are acquired differently.
iPhone uses a stereo matching method to obtain the depth, which is normalized to $0-255$ and saved as the inverse depth.
Huawei provides a low-resolution dense depth, i.e. $180\times240$ pixels, which is captured by a ToF sensor. 
Google Pixel uses a dual-pixel sensor, which has a very small baseline, to capture the depth.
The comparison is illustrated in Fig.~\ref{Fig: sensor complete cmp.}.
We can see that our method generalizes well to different real-world sparse data acquisition devices with a single trained model.
More examples are shown in Fig.~\ref{Fig: Pixel depth} and \ref{Fig: iPhone and Huawei}.

\noindent\textbf{Ablation of Synthetic Sparsity Patterns.}
This study aims to investigate the effectiveness of different simulated sparsity patterns during training.
We remove one of the proposed patterns during training and evaluate them on $3$ zero-shot datasets with different patterns. 
On NYU and ScanNet, we simulate the pattern `Sparse ToF' and `Unpaired FOV', while we use the provided sparse depth on Matterport3D. 
All models have been trained with the same number of epochs, and identical parameters other than the sparsity patterns.
Results are summarized in Table~\ref{Table: different sparse patterns.}. 
We observe that when missing the simulated sparse depth pattern, the performance on the most-related testing data will decrease.
For example, the model trained without the `Holes' pattern has worse performance than others on ScanNet and Matterport3D.
Therefore, we see how combining different sparsity patterns improves cross-domain generalization. 

\section{Discussion}
\noindent\textbf{Limitations.} 
Our method takes as input a depth map from the monocular depth estimation model as the guidance. 
While such single-image depth methods are improving, it is possible for the guidance map to have significant errors still.
This leads to adverse effects on depth completion.
Furthermore, although our method is robust to outliers and noise, we observe that the completion quality will decrease significantly in very sparse scenarios, e.g., where over $50\%$ of sparse depths are outliers.   

\noindent\textbf{Conclusion.}
In this paper, we analyze existing depth completion methods and show that they cannot generalize to different data domains and are sensitive to noise. We propose a simple yet robust system for depth completion.
Our method leverages a single image depth prior and is trained with diverse data augmentation. 
In addition, we redesign two depth completion benchmarks to better evaluate a depth completion method's generalization ability and its robustness to noise, different sparsity patterns, and diverse scenes.
We show that our approach achieves promising performance on our new benchmarks, works with a variety of sensor types and is robust to the sensor noise, with a single trained model. 
Our method provides a solution to high-quality mobile depth capture, which can improve downstream depth-related applications on mobile devices. 

\bibliographystyle{IEEEtran}
\bibliography{egbib}

\begin{thebibliography}{10}
\providecommand{\url}[1]{#1}
\csname url@samestyle\endcsname
\providecommand{\newblock}{\relax}
\providecommand{\bibinfo}[2]{#2}
\providecommand{\BIBentrySTDinterwordspacing}{\spaceskip=0pt\relax}
\providecommand{\BIBentryALTinterwordstretchfactor}{4}
\providecommand{\BIBentryALTinterwordspacing}{\spaceskip=\fontdimen2\font plus
\BIBentryALTinterwordstretchfactor\fontdimen3\font minus \fontdimen4\font\relax}
\providecommand{\BIBforeignlanguage}[2]{{%
\expandafter\ifx\csname l@#1\endcsname\relax
\typeout{** WARNING: IEEEtran.bst: No hyphenation pattern has been}%
\typeout{** loaded for the language `#1'. Using the pattern for}%
\typeout{** the default language instead.}%
\else
\language=\csname l@#1\endcsname
\fi
#2}}
\providecommand{\BIBdecl}{\relax}
\BIBdecl

\bibitem{shi2020pv}
S.~Shi, C.~Guo, L.~Jiang, Z.~Wang, J.~Shi, X.~Wang, and H.~Li, ``Pv-rcnn: Point-voxel feature set abstraction for 3d object detection,'' in \emph{IEEE Conference on Computer Vision and Pattern Recognition}, 2020, pp. 10\,529--10\,538.

\bibitem{wang2019pseudo}
Y.~Wang, W.-L. Chao, D.~Garg, B.~Hariharan, M.~Campbell, and K.~Q. Weinberger, ``Pseudo-lidar from visual depth estimation: Bridging the gap in 3d object detection for autonomous driving,'' in \emph{IEEE Conference on Computer Vision and Pattern Recognition}, 2019, pp. 8445--8453.

\bibitem{newcombe2011kinectfusion}
R.~A. Newcombe, S.~Izadi, O.~Hilliges, D.~Molyneaux, D.~Kim, A.~J. Davison, P.~Kohi, J.~Shotton, S.~Hodges, and A.~Fitzgibbon, ``Kinectfusion: Real-time dense surface mapping and tracking,'' in \emph{IEEE International Symposium on Mixed and Augmented Reality}.\hskip 1em plus 0.5em minus 0.4em\relax IEEE, 2011, pp. 127--136.

\bibitem{murORB2}
R.~Mur-Artal and J.~D. Tard\'os, ``{ORB-SLAM2}: an open-source {SLAM} system for monocular, stereo and {RGB-D} cameras,'' \emph{{IEEE} Transactions on Robotics}, vol.~33, no.~5, pp. 1255--1262, 2017.

\bibitem{xu2023frozenrecon}
G.~Xu, W.~Yin, H.~Chen, C.~Shen, K.~Cheng, and F.~Zhao, ``Frozenrecon: Pose-free 3d scene reconstruction with frozen depth models,'' in \emph{Proceedings of the IEEE/CVF International Conference on Computer Vision}, 2023, pp. 9310--9320.

\bibitem{schops2017multi}
T.~Schops, J.~L. Schonberger, S.~Galliani, T.~Sattler, K.~Schindler, M.~Pollefeys, and A.~Geiger, ``A multi-view stereo benchmark with high-resolution images and multi-camera videos,'' in \emph{IEEE Conference on Computer Vision and Pattern Recognition}, 2017, pp. 3260--3269.

\bibitem{yao2020blendedmvs}
Y.~Yao, Z.~Luo, S.~Li, J.~Zhang, Y.~Ren, L.~Zhou, T.~Fang, and L.~Quan, ``Blendedmvs: A large-scale dataset for generalized multi-view stereo networks,'' \emph{IEEE Conference on Computer Vision and Pattern Recognition}, 2020.

\bibitem{Zhang2019GANet}
F.~Zhang, V.~Prisacariu, R.~Yang, and P.~Torr, ``Ga-net: Guided aggregation net for end-to-end stereo matching,'' in \emph{IEEE Conference on Computer Vision and Pattern Recognition}, 2019, pp. 185--194.

\bibitem{zhang2018deepdepth}
Y.~Zhang and T.~Funkhouser, ``Deep depth completion of a single rgb-d image,'' \emph{IEEE Conference on Computer Vision and Pattern Recognition}, 2018.

\bibitem{senushkin2020decoder}
D.~Senushkin, I.~Belikov, and A.~Konushin, ``Decoder modulation for indoor depth completion,'' \emph{arXiv: Computing Research Repository}, p. 2005.08607, 2020.

\bibitem{huang2019indoor}
Y.-K. Huang, T.-H. Wu, Y.-C. Liu, and W.~H. Hsu, ``Indoor depth completion with boundary consistency and self-attention,'' in \emph{International Conference on Computer Vision Workshop}, 2019, pp. 0--0.

\bibitem{cheng2020cspn++}
X.~Cheng, P.~Wang, C.~Guan, and R.~Yang, ``Cspn++: Learning context and resource aware convolutional spatial propagation networks for depth completion,'' in \emph{Association for the Advancement of Artificial Intelligence}, vol.~34, 2020, pp. 10\,615--10\,622.

\bibitem{park2020non}
J.~Park, K.~Joo, Z.~Hu, C.-K. Liu, and I.-S. Kweon, ``Non-local spatial propagation network for depth completion,'' in \emph{European Conference on Computer Vision}.\hskip 1em plus 0.5em minus 0.4em\relax European Conference on Computer Vision, 2020.

\bibitem{xu2019depth}
Y.~Xu, X.~Zhu, J.~Shi, G.~Zhang, H.~Bao, and H.~Li, ``Depth completion from sparse lidar data with depth-normal constraints,'' in \emph{International Conference on Computer Vision}, 2019, pp. 2811--2820.

\bibitem{qiu2019deeplidar}
J.~Qiu, Z.~Cui, Y.~Zhang, X.~Zhang, S.~Liu, B.~Zeng, and M.~Pollefeys, ``Deeplidar: Deep surface normal guided depth prediction for outdoor scene from sparse lidar data and single color image,'' in \emph{IEEE Conference on Computer Vision and Pattern Recognition}, 2019, pp. 3313--3322.

\bibitem{cheng2019learning}
X.~Cheng, P.~Wang, and R.~Yang, ``Learning depth with convolutional spatial propagation network,'' \emph{IEEE Transactions on Pattern Analysis and Machine Intelligence}, 2019.

\bibitem{silberman2012indoor}
N.~Silberman, D.~Hoiem, P.~Kohli, and R.~Fergus, ``Indoor segmentation and support inference from rgbd images,'' in \emph{European Conference on Computer Vision}.\hskip 1em plus 0.5em minus 0.4em\relax Springer, 2012, pp. 746--760.

\bibitem{Uhrig2017THREEDV}
J.~Uhrig, N.~Schneider, L.~Schneider, U.~Franke, T.~Brox, and A.~Geiger, ``Sparsity invariant cnns,'' in \emph{International Conference on 3D Vision (3DV)}, 2017.

\bibitem{cheng2018depth}
X.~Cheng, P.~Wang, and R.~Yang, ``Depth estimation via affinity learned with convolutional spatial propagation network,'' in \emph{European Conference on Computer Vision}, 2018, pp. 103--119.

\bibitem{imran2021depth}
S.~Imran, X.~Liu, and D.~Morris, ``Depth completion with twin surface extrapolation at occlusion boundaries,'' \emph{IEEE Conference on Computer Vision and Pattern Recognition}, 2021.

\bibitem{tobin2017domain}
J.~Tobin, R.~Fong, A.~Ray, J.~Schneider, W.~Zaremba, and P.~Abbeel, ``Domain randomization for transferring deep neural networks from simulation to the real world,'' in \emph{{IEEE/RSJ} International Conference on Intelligent Robots and Systems}.\hskip 1em plus 0.5em minus 0.4em\relax IEEE, 2017, pp. 23--30.

\bibitem{tobin2018domain}
J.~Tobin, L.~Biewald, R.~Duan, M.~Andrychowicz, A.~Handa, V.~Kumar, B.~McGrew, A.~Ray, J.~Schneider, P.~Welinder \emph{et~al.}, ``Domain randomization and generative models for robotic grasping,'' in \emph{{IEEE/RSJ} International Conference on Intelligent Robots and Systems}.\hskip 1em plus 0.5em minus 0.4em\relax IEEE, 2018, pp. 3482--3489.

\bibitem{zakharov2019deceptionnet}
S.~Zakharov, W.~Kehl, and S.~Ilic, ``Deceptionnet: Network-driven domain randomization,'' in \emph{International Conference on Computer Vision}, 2019, pp. 532--541.

\bibitem{Wei2021CVPR}
W.~Yin, J.~Zhang, O.~Wang, S.~Niklaus, L.~Mai, S.~Chen, and C.~Shen, ``Learning to recover 3d scene shape from a single image,'' in \emph{IEEE Conference on Computer Vision and Pattern Recognition}, 2021.

\bibitem{Ranftl2020}
R.~Ranftl, K.~Lasinger, D.~Hafner, K.~Schindler, and V.~Koltun, ``Towards robust monocular depth estimation: Mixing datasets for zero-shot cross-dataset transfer,'' \emph{IEEE Transactions on Pattern Analysis and Machine Intelligence}, 2020.

\bibitem{Matterport3D}
A.~Chang, A.~Dai, T.~Funkhouser, M.~Halber, M.~Niessner, M.~Savva, S.~Song, A.~Zeng, and Y.~Zhang, ``Matterport3d: Learning from rgb-d data in indoor environments,'' \emph{International Conference on 3D Vision}, 2017.

\bibitem{dai2017scannet}
A.~Dai, A.~X. Chang, M.~Savva, M.~Halber, T.~Funkhouser, and M.~Nie{\ss}ner, ``Scannet: Richly-annotated 3d reconstructions of indoor scenes,'' in \emph{IEEE Conference on Computer Vision and Pattern Recognition}, 2017, pp. 5828--5839.

\bibitem{vasiljevic2019diode}
I.~Vasiljevic, N.~Kolkin, S.~Zhang, R.~Luo, H.~Wang, F.~Z. Dai, A.~F. Daniele, M.~Mostajabi, S.~Basart, M.~R. Walter \emph{et~al.}, ``Diode: A dense indoor and outdoor depth dataset,'' \emph{arXiv: Computing Research Repository}, p. 1908.00463, 2019.

\bibitem{schoenberger2016mvs}
J.~L. Sch\"{o}nberger, E.~Zheng, M.~Pollefeys, and J.-M. Frahm, ``Pixelwise view selection for unstructured multi-view stereo,'' in \emph{European Conference on Computer Vision}, 2016.

\bibitem{huynh2021boosting}
L.~Huynh, P.~Nguyen, J.~Matas, E.~Rahtu, and J.~Heikkila, ``Boosting monocular depth estimation with lightweight 3d point fusion,'' in \emph{International Conference on Computer Vision}, 2021, pp. 12\,767--12\,776.

\bibitem{chen2019learningJoint}
Y.~Chen, B.~Yang, M.~Liang, and R.~Urtasun, ``Learning joint 2d-3d representations for depth completion,'' in \emph{International Conference on Computer Vision}, 2019, pp. 10\,023--10\,032.

\bibitem{chen2019learning}
W.~Chen, S.~Qian, and J.~Deng, ``Learning single-image depth from videos using quality assessment networks,'' in \emph{IEEE Conference on Computer Vision and Pattern Recognition}, 2019, pp. 5604--5613.

\bibitem{yang2019dense}
Y.~Yang, A.~Wong, and S.~Soatto, ``Dense depth posterior (ddp) from single image and sparse range,'' in \emph{IEEE Conference on Computer Vision and Pattern Recognition}, 2019, pp. 3353--3362.

\bibitem{herrera2013depth}
D.~Herrera, J.~Kannala, J.~Heikkil{\"a} \emph{et~al.}, ``Depth map inpainting under a second-order smoothness prior,'' in \emph{Scandinavian Conference on Image Analysis}.\hskip 1em plus 0.5em minus 0.4em\relax Springer, 2013, pp. 555--566.

\bibitem{matsuo2015depth}
K.~Matsuo and Y.~Aoki, ``Depth image enhancement using local tangent plane approximations,'' in \emph{IEEE Conference on Computer Vision and Pattern Recognition}, 2015, pp. 3574--3583.

\bibitem{albishri2019cu}
A.~A. Albishri, S.~J.~H. Shah, and Y.~Lee, ``Cu-net: Cascaded u-net model for automated liver and lesion segmentation and summarization,'' in \emph{2019 IEEE International Conference on Bioinformatics and Biomedicine (BIBM)}.\hskip 1em plus 0.5em minus 0.4em\relax IEEE, 2019, pp. 1416--1423.

\bibitem{xu2022towards}
G.~Xu, W.~Yin, H.~Chen, C.~Shen, K.~Cheng, F.~Wu, and F.~Zhao, ``Towards 3d scene reconstruction from locally scale-aligned monocular video depth,'' \emph{arXiv preprint arXiv:2202.01470v3}, 2022.

\bibitem{eigen2014depth}
D.~Eigen, C.~Puhrsch, and R.~Fergus, ``Depth map prediction from a single image using a multi-scale deep network,'' in \emph{Advances in Neural Information Processing Systems}, 2014, pp. 2366--2374.

\bibitem{Yin2019enforcing}
W.~Yin, Y.~Liu, C.~Shen, and Y.~Yan, ``Enforcing geometric constraints of virtual normal for depth prediction,'' in \emph{International Conference on Computer Vision}, 2019.

\bibitem{liu2015learning}
F.~Liu, C.~Shen, G.~Lin, and I.~Reid, ``Learning depth from single monocular images using deep convolutional neural fields,'' \emph{IEEE Transactions on Pattern Analysis and Machine Intelligence}, vol.~38, no.~10, pp. 2024--2039, 2015.

\bibitem{xian2020structure}
K.~Xian, J.~Zhang, O.~Wang, L.~Mai, Z.~Lin, and Z.~Cao, ``Structure-guided ranking loss for single image depth prediction,'' in \emph{IEEE Conference on Computer Vision and Pattern Recognition}, 2020, pp. 611--620.

\bibitem{bian2021ijcv}
J.-W. Bian, H.~Zhan, N.~Wang, Z.~Li, L.~Zhang, C.~Shen, M.-M. Cheng, and I.~Reid, ``Unsupervised scale-consistent depth learning from video,'' \emph{International Journal of Computer Vision}, 2021.

\bibitem{monodepth2}
C.~Godard, O.~{Mac Aodha}, M.~Firman, and G.~J. Brostow, ``Digging into self-supervised monocular depth prediction,'' in \emph{International Conference on Computer Vision}, 2019.

\bibitem{yin2020diversedepth}
W.~Yin, X.~Wang, C.~Shen, Y.~Liu, Z.~Tian, S.~Xu, C.~Sun, and D.~Renyin, ``Diversedepth: Affine-invariant depth prediction using diverse data,'' \emph{arXiv: Computing Research Repository}, p. 2002.00569, 2020.

\bibitem{yin2020learning}
W.~Yin, J.~Zhang, O.~Wang, S.~Niklaus, L.~Mai, S.~Chen, and C.~Shen, ``Learning to recover 3d scene shape from a single image,'' \emph{IEEE Conference on Computer Vision and Pattern Recognition}, 2021.

\bibitem{ke2023repurposing}
B.~Ke, A.~Obukhov, S.~Huang, N.~Metzger, R.~C. Daudt, and K.~Schindler, ``Repurposing diffusion-based image generators for monocular depth estimation,'' in \emph{Proceedings of the IEEE/CVF Conference on Computer Vision and Pattern Recognition (CVPR)}, 2024.

\bibitem{xu2024diffusion}
G.~Xu, Y.~Ge, M.~Liu, C.~Fan, K.~Xie, Z.~Zhao, H.~Chen, and C.~Shen, ``Diffusion models trained with large data are transferable visual models,'' \emph{arXiv preprint arXiv:2403.06090}, 2024.

\bibitem{wang2022neuris}
J.~Wang, P.~Wang, X.~Long, C.~Theobalt, T.~Komura, L.~Liu, and W.~Wang, ``Neuris: Neural reconstruction of indoor scenes using normal priors,'' in \emph{European Conference on Computer Vision}.\hskip 1em plus 0.5em minus 0.4em\relax Springer, 2022, pp. 139--155.

\bibitem{yu2023improving}
X.~Yu, L.~Lu, J.~Rong, G.~Xu, and L.~Ou, ``Improving neural indoor surface reconstruction with mask-guided adaptive consistency constraints,'' \emph{arXiv preprint arXiv:2309.09739}, 2023.

\bibitem{ma2018sparse}
F.~Ma and S.~Karaman, ``Sparse-to-dense: Depth prediction from sparse depth samples and a single image,'' in \emph{International Conference on Robotics and Automation}.\hskip 1em plus 0.5em minus 0.4em\relax IEEE, 2018, pp. 4796--4803.

\bibitem{rosten2006machine}
E.~Rosten and T.~Drummond, ``Machine learning for high-speed corner detection,'' in \emph{European Conference on Computer Vision}.\hskip 1em plus 0.5em minus 0.4em\relax Springer, 2006, pp. 430--443.

\bibitem{esanet2020}
D.~Seichter, M.~K{\"o}hler, B.~Lewandowski, T.~Wengefeld, and H.-M. Gross, ``Efficient rgb-d semantic segmentation for indoor scene analysis,'' \emph{arXiv: Computing Research Repository}, 2020.

\bibitem{zamir2018taskonomy}
A.~Zamir, A.~Sax, , W.~Shen, L.~Guibas, J.~Malik, and S.~Savarese, ``Taskonomy: Disentangling task transfer learning,'' in \emph{IEEE Conference on Computer Vision and Pattern Recognition}.\hskip 1em plus 0.5em minus 0.4em\relax IEEE, 2018.

\bibitem{kim2018deep}
Y.~Kim, H.~Jung, D.~Min, and K.~Sohn, ``Deep monocular depth estimation via integration of global and local predictions,'' \emph{IEEE Transactions on Image Processing}, vol.~27, no.~8, pp. 4131--4144, 2018.

\bibitem{wang2020tartanair}
W.~Wang, D.~Zhu, X.~Wang, Y.~Hu, Y.~Qiu, C.~Wang, Y.~Hu, A.~Kapoor, and S.~Scherer, ``Tartanair: A dataset to push the limits of visual slam,'' \emph{arXiv: Computing Research Repository}, 2020.

\bibitem{wasenmuller2016comparison}
O.~Wasenm{\"u}ller and D.~Stricker, ``Comparison of kinect v1 and v2 depth images in terms of accuracy and precision,'' in \emph{Asian Conference on Computer Vision}.\hskip 1em plus 0.5em minus 0.4em\relax Springer, 2016, pp. 34--45.

\bibitem{he2016deep}
K.~He, X.~Zhang, S.~Ren, and J.~Sun, ``Deep residual learning for image recognition,'' in \emph{IEEE Conference on Computer Vision and Pattern Recognition}, 2016, pp. 770--778.

\bibitem{liu2017learning}
S.~Liu, S.~De~Mello, J.~Gu, G.~Zhong, M.-H. Yang, and J.~Kautz, ``Learning affinity via spatial propagation networks,'' \emph{arXiv: Computing Research Repository}, p. 1710.01020, 2017.

\bibitem{imran2019depth}
S.~Imran, Y.~Long, X.~Liu, and D.~Morris, ``Depth coefficients for depth completion,'' in \emph{IEEE Conference on Computer Vision and Pattern Recognition}.\hskip 1em plus 0.5em minus 0.4em\relax IEEE, 2019, pp. 12\,438--12\,447.

\bibitem{lee2021depth}
B.-U. Lee, K.~Lee, and I.~S. Kweon, ``Depth completion using plane-residual representation,'' in \emph{IEEE Conference on Computer Vision and Pattern Recognition}, 2021, pp. 13\,916--13\,925.

\bibitem{GargDualPixelsICCV2019}
R.~Garg, N.~Wadhwa, S.~Ansari, and J.~T. Barron, ``Learning single camera depth estimation using dual-pixels,'' \emph{International Conference on Computer Vision}, 2019.

\end{thebibliography}

\begin{figure}[h]
\centering
\includegraphics[width=0.3\textwidth]{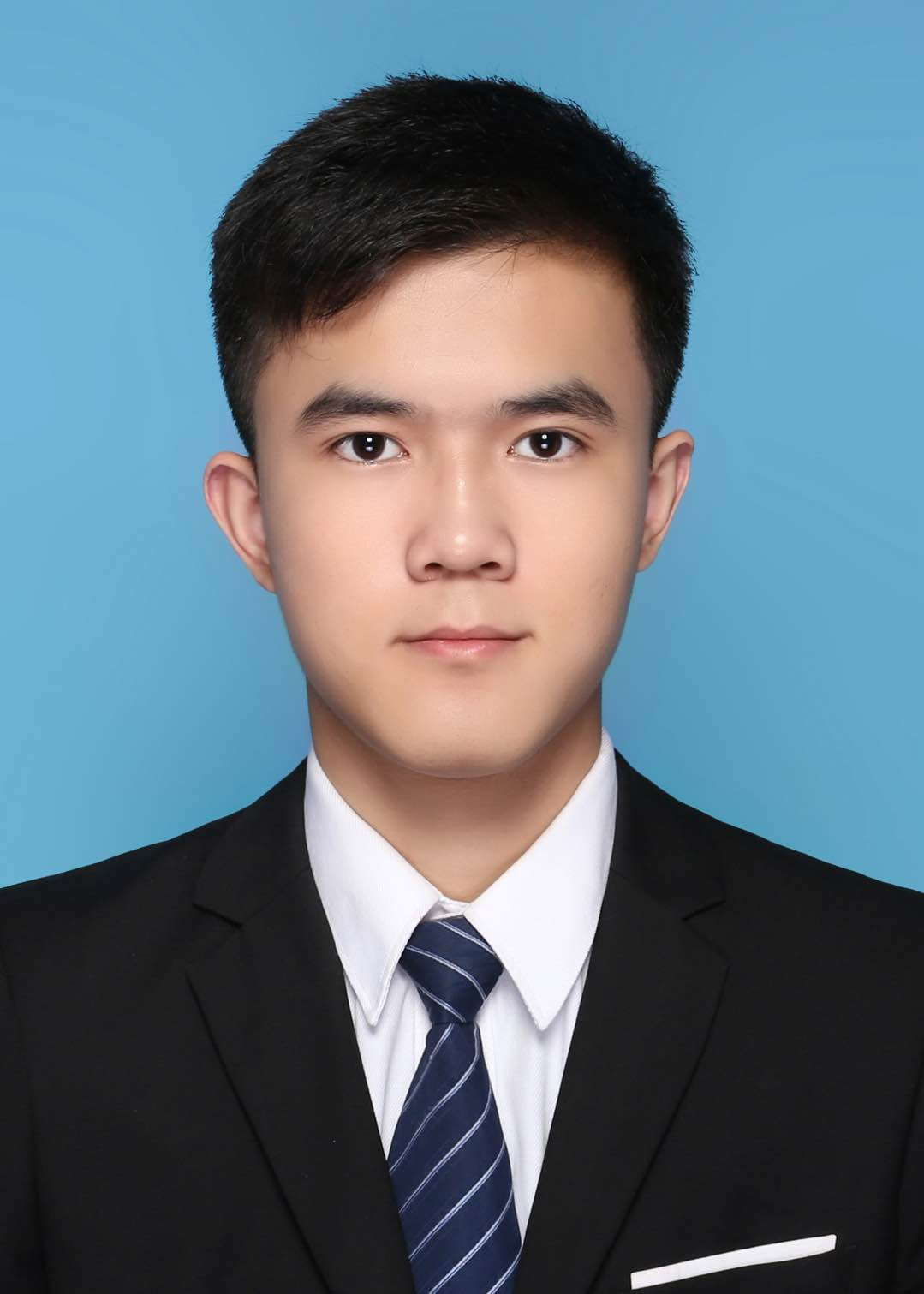}
\end{figure}

\noindent{\bf Guangkai Xu }\quad received the B. Eng. degree in the School of Automation Engineering from UESTC, China in 2020, and M. Eng. degree in the School of Information Science and Technology from USTC, China in 2023. Currently, he is a 1st-year Ph. D. student in the College of Computer Science and Technology in Zhejiang University.

His research interests include monocular depth estimation, 3D scene reconstruction and rendering, and visual-language models.

E-mail: guangkai.xu@gmail.com

ORCID iD: 0000-0001-9669-0381

\begin{figure}[h]
\centering
\includegraphics[width=0.3\textwidth]{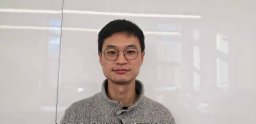}
\end{figure}

\noindent{\bf Wei Yin }\quad received the Ph.D. degree in computer science from University of Adelaide, Australia in 2022. 

His research interests include autonomous driving, 3D reconstruction. 

E-mail: yvanwy@outlook.com

\begin{figure}[h]
\centering
\includegraphics[width=0.3\textwidth]{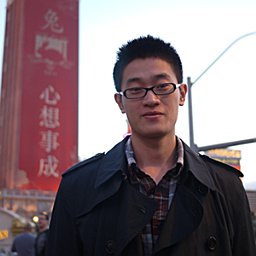}
\end{figure}

\noindent{\bf Jianming Zhang}\quad received the Ph.D. degree in computer vision with Prof. Stan Sclaroff at Boston University and got my PhD in 2016.

His research interests include deep learning, image processing, and intelligent systems.

E-mail: jianmzha@adobe.com

\begin{figure}[h]
\centering
\includegraphics[width=0.3\textwidth]{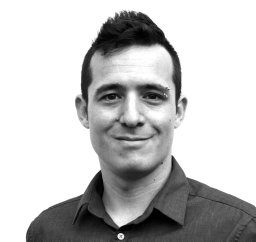}
\end{figure}

\noindent{\bf Oliver Wang}\quad is currently a senior staff research scientist at google research.

His research interests include image and video processing/editing, computer vision, machine learning, and photography.

E-mail: owang@adobe.com

\begin{figure}[H]
\centering
\includegraphics[width=0.3\textwidth]{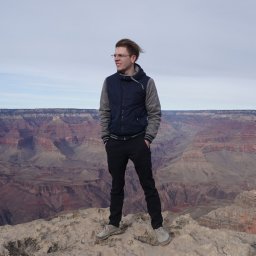}
\end{figure}

\noindent{\bf Simon Niklaus}\quad is a researcher at Adobe. He is a student of Feng Liu and is grateful for his internship at Adobe while working with Long Mai on the 3D Ken Burns project, and his internship at Google while working with Tianfan Xue on an undisclosed project within Marc Levoy‘s team, and his first years at Adobe when he was reporting to Oliver Wang.

His research interests include AI \& machine learning, computer vision, imaging \& video, graphics (2D \& 3D).

E-mail: sniklaus@pdx.edu

\noindent{\bf Simon Chen}\quad is currently a senior principal scientist at Adobe.

\begin{figure}[h]
\centering
\includegraphics[width=0.3\textwidth]{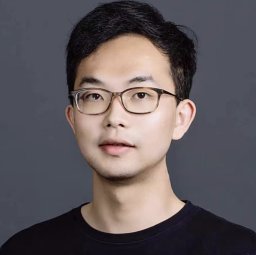}
\end{figure}

\noindent{\bf Jia-wang Bian}\quad 
received his B.Eng degree from Nankai University. After that, he did a research assistant job at the Singapore University of Technology and Design. He received his Ph.D. degree from the University of Adelaide. Also, Jiawang did research intern jobs in research institutes/companies, including the Advanced Digital Sciences Center, Tusimple, Amazon, and Facebook. He is currently a postdoctoral researcher at the University of Oxford. 

His research interests lie in the field of 3D Computer Vision.

E-mail: jiawang.bian@gmail.com

ORCID iD: 0000-0003-2046-3363

\end{document}